\documentclass[fleqn,11pt]{wlscirep}

 \twocolumn 

\newcommand{\edit}[1]{\textcolor{black}{#1}}

\usepackage{subcaption}
\usepackage{bm}
\usepackage{csquotes}
\usepackage{framed}
\usepackage{multicol}
\usepackage{amsmath}
\usepackage{changepage}
\graphicspath{ {images/} }

\def\reals{\mathbb{R}}
\newcommand{\capital}[1]{\bm{\mathrm{#1}}}
\title{DeepTag: inferring diagnoses from clinical notes in under-resourced medical domain}

\author[1,+]{Allen Nie}
\author[1,+]{Ashley Zehnder}
\author[2]{Rodney L. Page}
\author[1]{Arturo Lopez Pineda}
\author[1]{Manuel A. Rivas}
\author[1,3]{Carlos D. Bustamante}
\author[1,3, *]{James Zou}
\affil[1]{Department of Biomedical Data Science, Stanford University, Stanford, CA 94305, USA}
\affil[2]{Department of Clinical Sciences, Colorado State University, Fort Collins, CO 80523, USA}
\affil[3]{Chan-Zuckerberg Biohub, San Francisco, CA 94158, USA}

\affil[*]{jamesz@stanford.edu}

\affil[+]{these authors contributed equally to this work}

\begin{abstract}
Large scale veterinary clinical records can become a powerful resource for patient care and research. However clinicians lack the time and resource to annotate patient records with standard medical diagnostic codes and most veterinary visits are captured in free text notes. The lack of standard coding makes it challenging to use the clinical data to improve patient care. 
\edit{It is also a major impediment to cross-species translational research, which relies on the ability to accurately identify patient cohorts with specific diagnostic criteria in humans and animals}. \edit{In order to reduce the coding burden for veterinary clinical practice and aid translational research, we have developed a deep learning algorithm, DeepTag, which automatically infers diagnostic codes from veterinary free text notes.} DeepTag is trained on a newly curated dataset of 112,558 veterinary notes manually annotated by experts.
DeepTag extends multi-task LSTM with an improved hierarchical objective that captures the semantic structures between diseases. To foster human-machine collaboration, DeepTag also learns to abstain in examples when it is uncertain and defers them to human experts, resulting in improved performance. DeepTag accurately infers disease codes from free text even in challenging cross-hospital settings where the text comes from different clinical settings than the ones used for training. It enables automated disease annotation across a broad range of clinical diagnoses with minimal pre-processing. The technical framework in this work can be applied  in other medical domains that currently lack medical coding resources.

\end{abstract}
\begin{document}

\flushbottom
\maketitle
\thispagestyle{empty}

% \noindent Please note: Abbreviations should be introduced at the first mention in the main text – no abbreviations lists. Suggested structure of main text (not enforced) is provided below.

% \section*{Introduction}

% The Introduction section, of referenced text\cite{Figueredo:2009dg} expands on the background of the work (some overlap with the Abstract is acceptable). The introduction should not include subheadings.

\section*{Introduction}

% talk about the problem of lack of coding infrastructure across domains
% While a robust medical coding infrastructure exists in the US health care system for human medical records, this is not the case in veterinary medicine, which is under-resourced, suffering from a lack of coding infrastructure and standardized nomenclatures across medical institutions. 
While a robust medical coding infrastructure exists in the US healthcare system for human medical records, this is not the case in veterinary medicine, \edit{which is under-resourced in the sense that it lacks coding infrastructure and standardized nomenclatures across medical institutions. Most veterinary clinical notes are not coded with standard SNOMED-CT diagnosis~\cite{o2014approaches}.} 
% do we want to expand on the above point?
This hampers efforts at clinical research and public health monitoring. Due to the relative ease of obtaining large volumes of free-text veterinary clinical records for research (compared to similar volumes of human medical data) \edit{and the importance of turning these volumes of text into structured data to advance clinical research}, we investigated effective methods for building automatic coding systems for the veterinary records. 

%Importance of vet med and reason for current record problems
It is becoming increasingly accepted that spontaneous diseases in animals have important translational impact on the study of human disease for a variety of disciplines \cite{kol2015companion}. Beyond the study of zoonotic diseases, which represent 60-70\% of all emerging diseases \footnote{\url{https://wwwnc.cdc.gov/eid/page/zoonoses-2018}}, non-infectious diseases, like cancer, have become increasingly studied in companion animals as a way to mitigate some of the problems with rodent models of disease \cite{leblanc2015defining}. \edit{Additionally, spontaneous models of disease in companion animals are being used in drug development pipelines as these models more closely resemble the ``real world'' clinical settings of diseases than genetically altered mouse models~\cite{grimm2016bark,klinck2017translational,baraban2014new,hernandez2018naturally}. %Veterinary clinical records are currently not tagged with disease codes such as SNOMED-CT, and this is a major impediment to leveraging veterinary data to study diseases. DeepTag is designed to address this need.
} However, when it comes to identifying clinical cohorts of veterinary patients on a large scale for clinical research, there are several problems.  One of the first is that veterinary clinical visits rarely have diagnostic codes applied to them, either by clinicians or medical coders.  There is no substantial third party payer system and no HealthIT act that applies to veterinary medicine, so there are few incentives for clinicians or hospitals to annotate their records for diseases to be able to identify patients by diagnosis.  Billing codes are largely institution-specific and rarely applicable across institutions, unless hospitals are under the same management structure and records system. Some large corporate practice groups have their own internal clinical coding structures, but that data is rarely made available for outside researchers. A small number ($< 5$) academic veterinary centers (of a total of 30 veterinary schools in the US)\footnote{\url{http://aavmc.org/}} employ dedicated medical coding staff that apply disease codes to clinical records so these records can be identified for clinical faculty for research purposes. How best to utilize this rare, well-annotated, veterinary clinical data for the development of tools that can help organize the remaining seqments of the veterinary medical domain is an open area of  research. 

\edit{In this paper, we develop DeepTag, a system to automatically code veterinary clinical notes. DeepTag takes free-form veterinary note as input and infers clinical diagnosis from the note. The inferred diagnosis is in the form of SNOMED-CT codes. We trained DeepTag on a large set of 112,558 veterinary notes, and each note is expert labeled with a set of SNOMED-CT codes. DeepTag is a bidirectional long-short term memory network (BLSTM) augmented with a hierarchical training objective that captures similarities between the diagnosis codes. We evaluated DeepTag's performance on challenging cross-hospital coding tasks.}

% rewrite Related Work section

\paragraph{Related work}
% 1. Overall NLP progress
Natural language processing (NLP)  techniques have improved from leveraging discrete patterns such as n-grams \cite{jurafsky2014speech} to continuous learning algorithms like Long-short-term Memory Networks (LSTMs) \cite{hochreiter1997long}. This strategy has proven to be extremely successful when a sizable amount of data can be acquired. \edit{Combined with advances in optimization and classification algorithms, the field has developed algorithms that match or exceed human performance in traditionally difficult tasks in multiple domains \cite{goldberg2017neural}.}

% 2. Clinical NLP progress / problems
\edit{Analyzing free text such as diagnostic reports and clinical notes has been a central focus of clinical natural language processing~\cite{velupillai2015recent}. Most of the previous research has focused on the human healthcare systems. Examples include using NLP tools to improve pneumonia screening in the emergency department, assisting in adenoma detection, assisting and simplifying hospital processes by identifying billing codes from clinical notes~\cite{demner2016aspiring}. Pivovarov et al. have conducted experiments to discover phenotypes and diseases using an unsupervised method on a broad set of heterogeneous data~\cite{pivovarov2015learning}.}
% can add: due to lack of available data, previous exploration has been focusing on human medicine domain.

% 3. Why these problems don't exist in Veterinary medicine (because something is better than nothing)
% we argue that NLP tools can be more impactful in the domain of veterinary medicine.
\edit{In the domain of veterinary medicine, millions of clinical summaries are stored as electronic health records (EHR) in various hospitals and clinics. Unlike human discharge summaries that have been assigned with billing codes (ICD-9/ICD-10 codes), veterinary summaries exist primarily as free text. This makes it challenging to perform  systematic analysis  such as disease prevalence studies, analysis of adverse drug effects, therapeutic efficacy or outcome analysis. Veterinary domain is very favorable for an NLP system that can convert large amount of  free-text notes into structured information. Such a system would benefit the veterinary community in a substantial way and can be deployed in multiple clinical settings. Veterinary medicine is a domain where clinical NLP tools can have a substantial impact in practice and be integrated into daily use.}
% bird flu, swine flu, AIDs (all spread from animal to people)

% 4. Talk about NER and automated coding 
\edit{Identifying a set of conditions/diseases from clinical notes has been actively studied~\cite{demner2016aspiring,lipton2015learning}. Currently, the task of transforming free text into structured information primarily relies on two approaches: named entity recognition (NER) and automated coding. DeepTag is designed to perform automated coding rather than NER. NER requires annotation on the word level, where each word is associated with one of a few types. In the ShARe task \cite{pradhan2014evaluating}, the importance is placed on identifying disease span and then normalizing into standard terminology in SNOMED-CT or UMLS (Unified Medical Language System). In other works, the focus has been on tagging each word with a specific type: adverse drug effect, severity, drug name, etc~\cite{jagannatha2016bidirectional}. Annotating on word level is expensive, and most corpora contain only a couple of hundreds or thousands of clinical notes. Even though early shared task in this domain has proven to be successful  ~\cite{pradhan2014semeval,elhadad2015semeval}, it is still difficult to curate a large dataset in this manner.}

% add a new citation (the one with hierarchical system); how we leverage hierarchy slightly differently than this other work.
\edit{Automated coding on the other hand takes the entire free text as input, and infers a set of labels that are used to code the entire work. Most discharge summaries in human hospitals have billing codes assigned. Baumel et al. proposed a text processing model for automated coding that processes each sentence first and then processes the encoded hidden states for the entire document~\cite{baumel2017multi}.This multi-level approach is especially suitable for longer texts, and the method was applied to the MIMIC data, where each document is on average five times longer than the veterinary notes from CSU. Rajkomar et al. used deep learning methods to process the entire EHR and make clinical predictions for a wide range of problems including automated coding~\cite{rajkomar2018scalable}. In their work, they compared three deep learning models: LSTM, time-aware feedforward neural network (TANN), and boosted time-based decision stumps. In this work, we use a new hierarchical training objective which is designed to capture the similarities among the SNOMED-CT codes. This hierarchical objective is complementary to these previous approaches in the sense that the hierarchical objective can be used on top of any architecture. Our cross-hospital evaluations also extend what is typically done in literature. Even though Rajkomar et al. had data from two hospitals, they did not investigate the performance of the model when trained on one hospital but evaluated on the other. In our work, due to the lack of coded clinical notes in the veterinary community beyond a few academic hospitals, it is especially salient for us to evaluate the model’s ability to generalize across hospitals.}

\edit{Our work is also related to the work of Kavuluru et al., who experimented with different training strategies and compared which strategy is the best for automated coding~\cite{kavuluru2015empirical}, and Subotin et al., who improved upon direct label probability estimation and used a conditional probability estimator to fine-tune the label probability~\cite{subotin2016method}. Perotte et al. also investigated possible methods to leverage the hierarchical structure of disease labels by using an SVM algorithm on each level of the ICD-9 hierarchy tree~\cite{perotte2013diagnosis}.}

% Both Baumel et al. and Rajkomar et al. proposed new text processing models leveraging either hierarchical attention, or building time-aware feed-forward attention model~\cite{baumel2017multi,rajkomar2018scalable}. In this work, our contribution is orthogonal to their innovations in the sense that we seek to augment the loss function so that it can be applied to any text processing model. By experimenting the standard state-of-the-art model, we aim to encourage more people to utilize our innovation. Our work is more similar to the early work of Kavuluru et al., who experimented with different training strategies and compared which strategy is the best for the automated coding setting~\cite{kavuluru2015empirical}, and Subotin et al., who improved upon direct label probability estimation and used a conditional probability estimator to finetune the label probability~\cite{subotin2016method}. Perotte et al. also investigated possible methods to leverage the hierarchical structure of disease labels by using a SVM algorithm on each level of the ICD-9 hierarchy tree~\cite{perotte2013diagnosis}. This has the problem of introducing cascading errors at test time. Our clustering strategy produces the same set of labels as any flat prediction algorithm and only enforce semantic similarities between diseases through augmenting the loss function.

\begin{figure*}[h!]
    \begin{subfigure}[b]{\textwidth}
        \centering
        \includegraphics[scale=0.4]{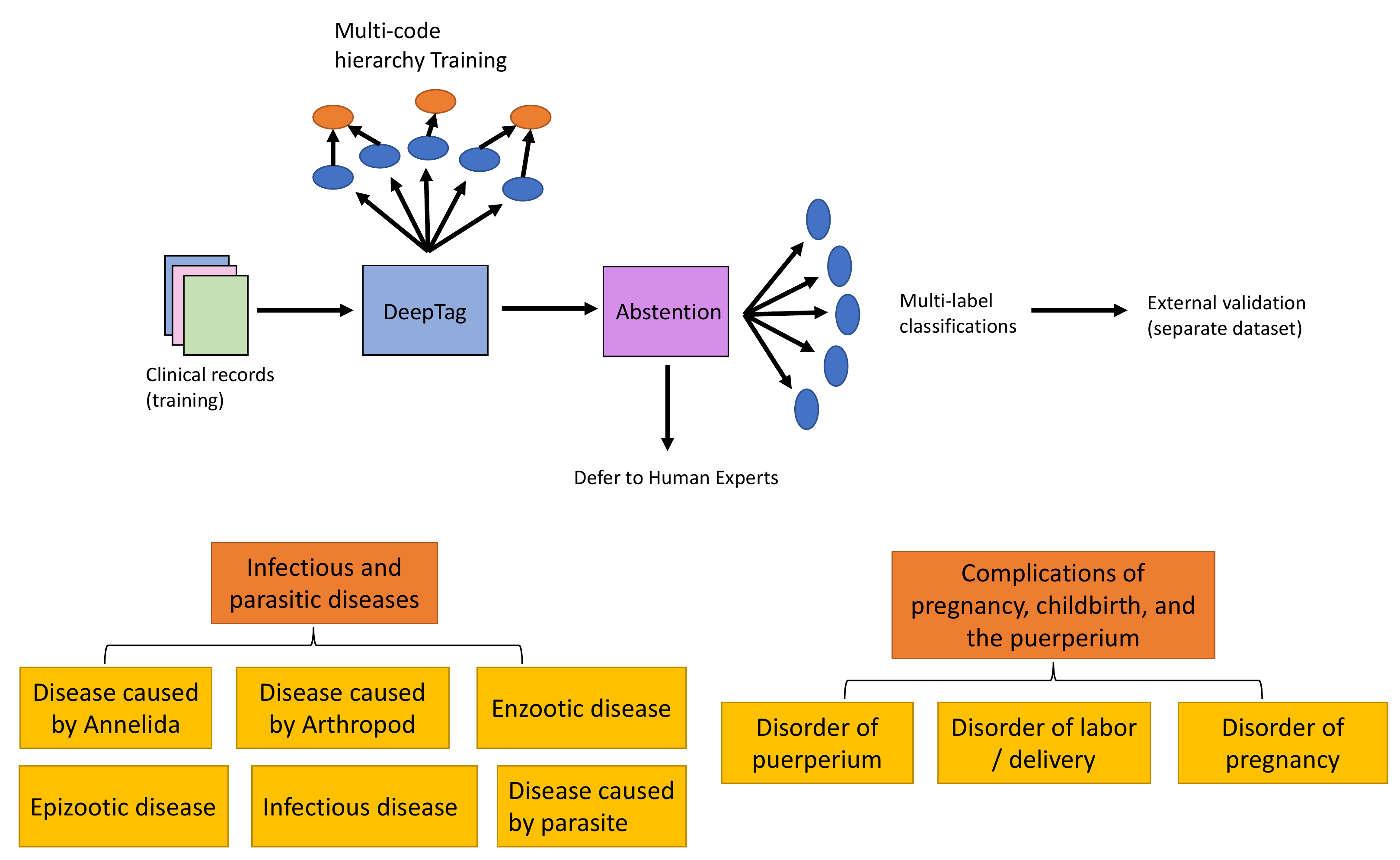}
        \caption{}
        \label{fig:work-schematic}
    \end{subfigure}
    \hfill
    \begin{subfigure}[b]{\textwidth}
        \footnotesize
        \textbf{CSU clinical note example}
        \begin{framed}
        Jem is a 10 year old male castrated hound mix that was presented for continuation of chemotherapy for previously diagnosed B-cell multicentric lymphoma. Jem was started on CHOP chemotherapy last week and has been doing very well since receiving doxorubicin. The owners have noted his lymph nodes have gotten much smaller. He has some loose stool, yet improved with metronidazole. Current medications include prednisolone. Assessment: Jem is in a strong partial remission based on today's physical exam. He is also doing very well since starting chemotherapy. A CBC today was unremarkable and adequate for chemotherapy. She was dispensed oral cyclophosphamide and furosemide that the owners were instructed to give at home.
        \vspace{0.1in}
        
        \textbf{Expert annotated diseases}: Disorder of hematopoietic cell proliferation, Neoplasm and/or hamartoma
        % Disease codes: Disorder of hematopoietic cell proliferation(28), Neoplasm and/or hamartoma(40)
        \end{framed}
        
        \textbf{PP clinical note example}
        \begin{framed}
        Likely ear infection  shaking head  now swollen  drooping ear otherwise doing well  amublating well- had RF carpal arthrodesis at UCD. wt:  95.3 lbs.    Ears/Eyes/Nose/Throat: Clear OU/ brown yeasty debris AU errythema AU no fb/tm;s intact AU mod aural hematoma AD  Cardiovascular:   No murmur/arrhythmia.  Femoral pulses strong and synchronous. HR:84  Respiratory:  Lungs sound clear bilaterally  no crackles or wheezes. Eupneic.  RR:20  Lymph nodes:   No palpable peripheral lymphadenopathy  Oral Exam:  mm pink and moist.   CRT $<$ 2 sec  Musculoskeletal:  No lameness noted. Arthrodesis carpal joint RF  thickened stifle LH  ambulating well  Nervous System:   Appropriate mentation.  No overt neurologic deficits  Integument:  Full haircoat. Adequate skin turgor. otitis externa AU  aural hematoma AD. Ear cytology -  ++cocci  yeast   AU.  Cleaned  with epiotic. Rx Tresaderm BID x  14 days.  applied first dose    Rx temarilP taper    Skin prep medial AD over hematoma.  then 19g butterfly needle attached to syringe  drained 10ml bloody fluid  held pressure with guaze  no more bleeding        CE: Recheck in 14d to ensure infection cleared  discussed aural hematomas  options for tx  sx. may recur. 
        \vspace{0.1in}
    
        \textbf{Expert annotated diseases}: Infectious disease, Disorder of the integument, Disorder of the auditory system
        \end{framed}
        \caption{}
        \label{fig:csu-pp-text-examples}
    \end{subfigure}
    \caption{{\bf System workflow and clinical note examples.} Figure (a) shows  the workflow of DeepTag  with abstention. Then we show two example meta-categores corresponding to two subsets of the 42 SNOMED-CT codes. Figure (b) shows two example notes from the CSU and PP datasets. The highlighted text shows the supporting evidence human curators use to assign disease tags to these documents.  }
    \label{fig:figure1}
\end{figure*}

\edit{Cross-hospital generalization is a significant challenge in the veterinary coding setting. Most veterinary clinics currently do not apply diagnosis codes to their notes~\cite{o2014approaches}. Therefore our training data can only come from a handful of university-based regional referral centers that manually code their free text notes. The task is to train a model on such data and deploy for thousands of private hospitals and clinics. University-based centers and private hospitals and clinics have substantial variation in the writing style, the patient population, and the distribution of diseases (Figure~\ref{fig:csu-pp-text-examples}). For example, the training dataset we have used in this work comes from a university-based hospital with a high-volume referral oncology service, but typical local hospitals might face more dermatologic or gastrointestinal cases.}

% \james{Remove the bold words to avoid confusion with NER.}
% \james{Label the top and bottom figures of Fig 1 as (a) and (b) and refer to it as (a) and (b) in the caption.}

\section*{Results}
DeepTag takes clinician's notes as input and predicts a set of SNOMED-CT disease codes. SNOMED-CT is a comprehensive and precise clinical health terminology managed by the International Health Terminology Standards Development Organization. DeepTag is a bi-directional long short-term memory (LSTM) neural network with a new hierarchical learning objective designed to capture similarities between the SNOMED-CT codes. 
%\edit{We leverage the fact that we can find correspondence between SNOMED-CT disease codes and ICD-9 top-level codes (as seen in Figure~\ref{fig:work-schematic}, and we provide the complete correspondence list in supplementary material) to develop a novel hierarchical objective function which improved the generalization performance of DeepTag.}

% We leverage the fact that the SNOMED-CT codes are structured and developed a novel hierarchical objective function which improved the generalization performance of DeepTag.

DeepTag is trained on 112,558 annotated veterinary notes from the Colorodo State University of Veterinary Medicine and Biomedical Sciences (CSU) for research purposes. Each of these notes is a free text description of a patient visit, and is manually labeled with at least one, but on average eight, SNOMED-CT codes by experts. 
% ANIE: Hi Dr. Zou, the highest number of codes on a single document is 112, and I don't think "between 1 and 112 codes" is informative here...
 %SNOMED-CT code is designed hierarchically, with lower level codes being more specific and higher level codes being more broad. in order to make DeepTag useful in different usage scenarios, where the distribution of lower level codes can change completely, mapped lower level specific SNOMED-CT codes to 95 SNOMED-CT disease/disorder codes, which reside on a higher level of the SNOMED-CT hierarchy.
When the coder-applied disease codes that are mapped up to the children of parent note \emph{Disease (disorder)} (ConceptID: 64572001), there are 41 SNOMED-CT top-level disease codes present in the CSU dataset. In addition, we map every non-disease related code to an extra spurious code. In total, DeepTag learns to tag a clinical note with a subset of 42 codes. 

We evaluate DeepTag on two different datasets. One consists of 5,628 randomly sampled non-overlapping documents from the same CSU dataset that the system is trained on. The other dataset contains 586 documents and are collected from a private practice (PP) located in northern California. Each of the these document is also manually annotated with the appropriate SNOMED-CT codes by human experts. We refer to this dataset as the PP dataset. 
% \edit{``Out-of-domain'' in natural language processing has been defined as having a significant shift in vocabulary, format, and text style \cite{li2012literature}.}
We regard the PP dataset as a ``out-of-domain'' dataset due to its substantial difference with regard to writing style and institution type compared to the CSU dataset~\cite{li2012literature}.

\subsection*{Tagging performance}

% First Q: why is it doing better in CSU?
% why neoplasm does much better, compare to cell proliferation; why is disorder of 
% why are some categories doing so well (regardless of frequency)
% what are the diseases that coders are actually putting in (look at extreme cases -- Disorder of digestive system, why is it hard? Too many diseases lumped into one category? Any other features about text that could make the category more difficult) (quantitative analysis)
% another extreme is disease with very few examples that are still doing well (is it because it has very few diseases lumped into them?)

% for discussion, just talk about two extremes for qualitative analysis 

% Second Q: why is there a drop in PP?
% match things that are not in the standard English dictionary (30,000 English words) (match in CSV and match in PP, and say the ratio)

% also talk about why precision didn't drop on rarer diseases, while recall really dropped
% PP we lose mostly in recall because of vocabulary (out of vocabulary that's causing recall to drop) (keywords we know are still good)

% Correlation goes before the error analysis

\begin{table*}[!t]
\centering
\footnotesize
\begin{adjustwidth}{-0.5in}{0in}
\caption{\textbf{Report of DeepTag performance on CSU test data and PP data}}
\begin{tabular}{c | c c c c c c | c c c c c c }
\toprule
 & \multicolumn{6}{c|}{CSU} & \multicolumn{6}{c}{PP (\edit{Cross-hospital})}  \\
Disease & \edit{N} & Prec & Rec & $F_1$ & Accu & Sub & \edit{N} & Prec & Rec & $F_1$ & Accu  & Sub \\
\midrule
Autoimmune disease  & 1280 & 94.0 & 72.3 & 81.4 & 99.6 & 11 & 1 & 60.0 & 25.0 & 34.7 & 99.6 & 1(1) \\ 
% Disorder of hemostatic system  & 1660 & 80.9 & 54.0 & 64.5 & 99.1 & 13 & 1 & 50.0 & 25.0 & 31.3 & 99.4 & 1(0) \\ 
% Disorder of cellular component of blood  & 2263 & 74.0 & 54.0 & 61.6 & 98.6 & 33 & 7 & 18.3 & 7.3 & 10.4 & 98.1 & 5(0) \\ 
Congenital disease  & 3345 & 72.9 & 35.9 & 47.3 & 97.8 & 224 & 17 & 58.0 & 5.3 & 9.2 & 97.1 & 8(6) \\ 
Propensity to adverse reactions  & 5105 & 89.1 & 70.2 & 78.1 & 98.2 & 8 & 43 & 83.6 & 13.2 & 21.3 & 93.0 & 7(2) \\ 
Metabolic disease  & 5265 & 68.9 & 55.4 & 61.0 & 96.9 & 82 & 26 & 75.8 & 51.1 & 59.8 & 96.5 & 12(9) \\ 
Disorder of auditory system  & 5393 & 81.0 & 66.2 & 72.8 & 97.7 & 67 & 64 & 76.8 & 69.4 & 72.4 & 95.0 & 12(6) \\ 
Hypersensitivity condition  & 6871 & 85.7 & 74.6 & 79.5 & 97.7 & 31 & 50 & 72.7 & 20.6 & 30.5 & 92.1 & 11(4) \\ 
Disorder of endocrine system  & 7009 & 79.2 & 66.7 & 72.2 & 96.9 & 84 & 46 & 71.5 & 28.2 & 40.1 & 92.6 & 8(8) \\ 
Disorder of hematopoietic cell proliferation  & 7294 & 95.1 & 87.4 & 91.0 & 98.9 & 22 & 16 & 68.6 & 31.0 & 41.1 & 97.5 & 6(1) \\ 
Disorder of nervous system  & 7488 & 76.1 & 63.8 & 69.2 & 96.4 & 243 & 27 & 68.0 & 31.7 & 41.6 & 94.9 & 19(14) \\ 
Disorder of cardiovascular system  & 8733 & 79.3 & 62.5 & 69.7 & 95.7 & 351 & 53 & 48.1 & 54.7 & 49.9 & 89.8 & 30(24) \\ 
Disorder of the genitourinary system  & 8892 & 77.7 & 62.6 & 69.3 & 95.7 & 317 & 44 & 59.9 & 40.3 & 47.2 & 92.3 & 19(12) \\ 
Traumatic AND/OR non-traumatic injury  & 9027 & 72.8 & 57.2 & 63.5 & 94.8 & 536 & 19 & 43.8 & 12.0 & 18.3 & 96.2 & 13(8) \\ 
Visual system disorder  & 10139 & 84.3 & 81.1 & 82.6 & 96.9 & 413 & 62 & 68.6 & 63.9 & 65.7 & 93.2 & 39(34) \\ 
Infectious disease  & 11304 & 71.2 & 53.7 & 60.8 & 92.9 & 260 & 88 & 56.3 & 21.9 & 30.2 & 86.7 & 20(10) \\ 
Disorder of respiratory system  & 11322 & 79.5 & 65.5 & 71.8 & 95.2 & 274 & 27 & 40.2 & 41.3 & 38.4 & 94.0 & 16(14) \\ 
Disorder of connective tissue  & 17477 & 75.4 & 67.0 & 70.7 & 91.3 & 567 & 24 & 38.4 & 31.2 & 33.8 & 94.3 & 15(11) \\ 
Disorder of musculoskeletal system  & 20060 & 77.0 & 73.4 & 74.8 & 91.1 & 670 & 56 & 66.4 & 45.4 & 53.2 & 91.5 & 31(19) \\ 
Disorder of integument  & 21052 & 84.2 & 71.6 & 77.3 & 92.3 & 360 & 156 & 55.7 & 58.5 & 56.8 & 80.6 & 58(32) \\ 
Disorder of digestive system  & 22589 & 76.8 & 67.1 & 71.5 & 89.7 & 694 & 195 & 54.7 & 48.4 & 50.2 & 72.7 & 47(36) \\ 
Neoplasm and/or hamartoma  & 36108 & 92.2 & 88.9 & 90.5 & 93.9 & 749 & 59 & 27.5 & 77.3 & 40.0 & 76.2 & 18(7) \\ 
\bottomrule
\end{tabular}
\begin{flushleft} This table reports the DeepTag's performance (precision, recall, $F_1$ and accuracy) for the 21 most frequent disease categories (from a total of 42 categories). $N$ indicates the total number of examples in the dataset. Sub indicates the number of specific disease codes that are present in the dataset that are binned into one of the disease level codes. For the PP dataset, the Sub number in parentheses indicate the number of subtypes that are also present in CSU dataset. 
\end{flushleft}
 \label{tab:detailed-csu-pp}
\end{adjustwidth}
\end{table*}

We present DeepTag's performance on the CSU and PP test data in Table~\ref{tab:detailed-csu-pp}. To save space, we display the 21 most frequent disease codes in Table~\ref{tab:detailed-csu-pp}. Each SNOMED-CT code corresponds to one disease category. For each category, we report the number of training examples in the category ($N$), the scores for precision, recall, $F_1$, accuracy, and the number of disease subtypes in this category. While DeepTag achieves reasonable $F_1$ scores overall, its performance is quite heterogeneous in different categories. Moreover the performance decreases when DeepTag is applied to the out-of-domain PP test data.

%Analysis of DeepTag performance needs to account for the multi-label aspect of the prediction task. 
We identify two factors that substantially impact DeepTag's performance: 1) the number of training examples that are tagged with the given disease label; \edit{2) the number of subtypes, where a subtype is a SNOMED-CT code applied to either dataset that is lower in the SNOMED-CT hierarchy than the top-level disease category codes DeepTag is predicting.} We use the number of subtypes as a proxy for the diversity and specificity of the clinical text descriptions.  Thus, a higher number of subtypes is used here as a proxy for a wider spectrum of diseases.

% 2) the number of subtypes, which are SNOMED-CT codes that are actually applied to the CSU dataset that are lower in the SNOMED-CT hierarchy compared to the disease-level codes DeepTag is predicting

%For example, \emph{Neoplasm and/or hamartoma} encapsulates many different histologic types and be categorized as benign, malignant, or unknown, thus resulting in many different lower-level codes (749 codes) being mapped into the same top-level disease code.

%We then look into the problem of performance decrease when we apply DeepTag to an out-of-domain dataset, PP. The primary contributing factor is that the underlying text in PP is stylistically and functionally different compared to text used for training in the CSU data, which can be seen in Figure~\ref{fig:csu-pp-text-examples}. This results in a sharp drop in terms of both exact match ratio (EM) and per-label $F_1$ score across all labels. We account the domain difference by computing the non-common word vocabulary match between the CSU and PP domain.

\begin{figure}[!h]
\centering
\includegraphics[scale=0.4]{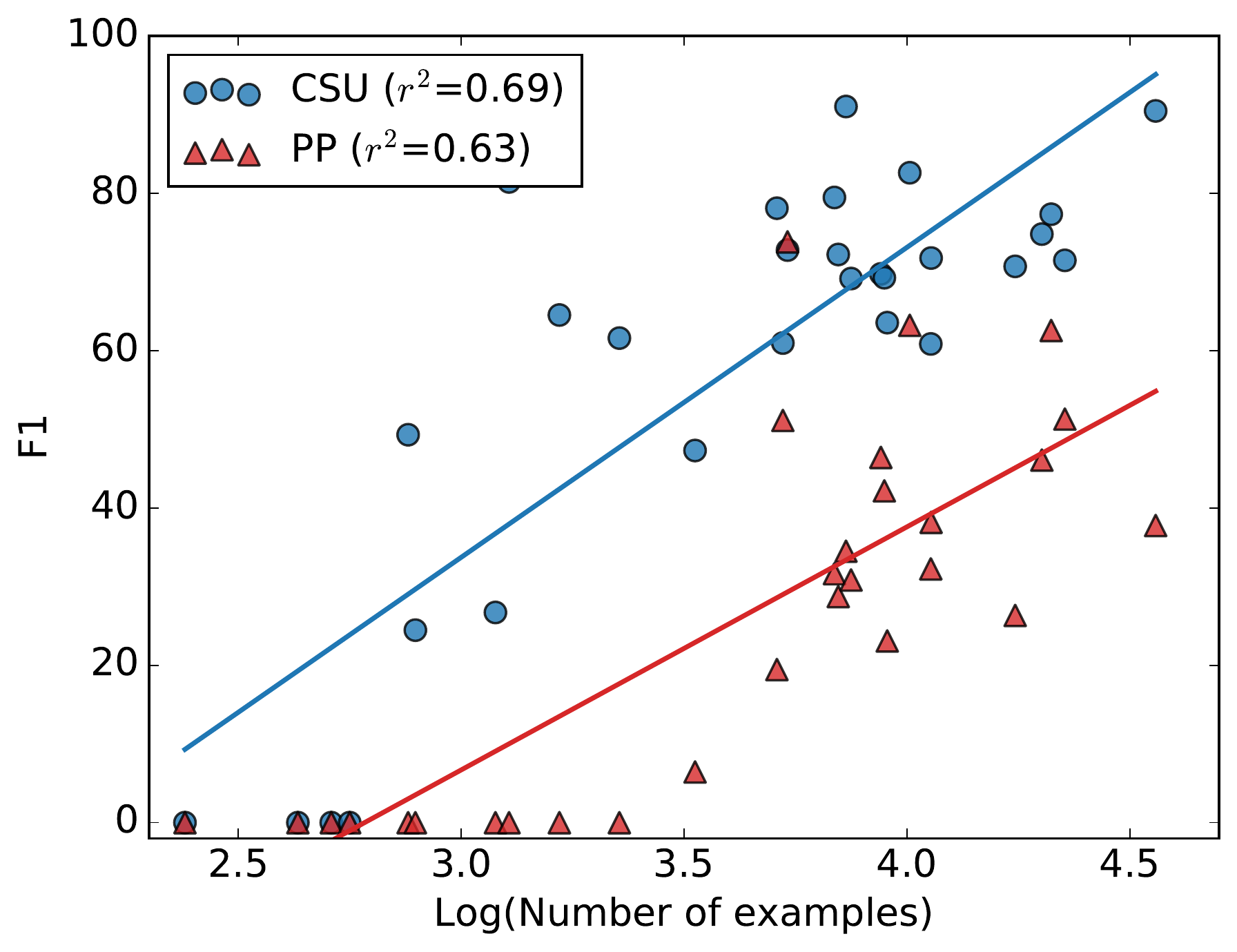}
\caption{{\bf Per-label $F_1$ score plotted with log of number of examples in the training dataset.} Results shown here are from the DeepTag model. Each point represents a label, its corresponding number of training examples in CSU, and the per-label $F_1$ score from the DeepTag model.
}
\label{fig:example-f1-scatterplot}
\end{figure}

\paragraph{Performance improves with more training examples.}
We first note that DeepTag works relatively well when the number of training examples for each label is abundant. We generate a scatter plot to capture the correlation between number of examples in the in CSU dataset and the label's $F_1$ score evaluated on the CSU test set. We also plot the $F_1$ score for the label evaluated on PP dataset and its number of training examples on CSU dataset. 
% We investigate how strong does number of training examples correlate with in-domain test set performance (CSU) and whether it transfers well to out-of-domain performance (PP).
% We show this finding in Figure~\ref{fig:example-f1-scatterplot}.

For the CSU dataset, we observe an almost linear relationship between the log number of examples and the $F_1$ score in Figure~\ref{fig:example-f1-scatterplot}. 
%a few exceptions, which we will analyze in more detail. 
We observe a similar pattern when evaluating on PP dataset, thought the correlation is weaker and the pattern is less linear. This is due to the out-of-domain nature of PP, which we investigate in depth below. % but having more training examples in CSU is not a very strong indicator on how well the system performs on PP, and the correlation is possibly driven by labels that have few CSU training examples and get zero $F_1$ scores. %We report $r^2$ from an ordinary least square (OLS) analysis and the coefficient is positive with $p < .001$.

% Largely speaking, more training examples in CSU will correspond to better performance in that category in PP, though the relation is weaker and more outliers can be observed. We report $r^2$ from ordinary linear regression (OLS) analysis between log(number of training examples) and F1 in Figure~\ref{fig:example-f1-scatterplot}.

\begin{table*}[!h]
% \begin{adjustwidth}{-2.25in}{0in} % Comment out/remove adjustwidth environment if table fits in text column.
\centering
\footnotesize
\caption{\textbf{Evaluation of  trained classifiers on the CSU and PP data}}
\begin{tabular}{c | c | c c | c c | c c }
\toprule
Model & \multicolumn{1}{c|}{EM} & \multicolumn{2}{c|}{Precision} & \multicolumn{2}{c|}{Recall} & \multicolumn{2}{c}{$F_1$} \\
& & unwgt & wgt & unwgt & wgt & unwgt & wgt \\
\midrule
& \multicolumn{7}{c}{CSU data} \\
\midrule
LSTM & 47.4 & 76.6 & 85.9 & 59.3 & 78.7 & 65.3 & 81.7 \\ % & $\pm 0.5$
BLSTM & 48.2 & 76.1 & 86.0 & 57.6 & 79.4 & 63.5 & 82.2 \\ % & $\pm 0.9$
DeepTag-M & 48.6 & 76.8 & 86.3 & 58.7 & 79.6 & 64.6 & 82.4 \\ % & $\pm 0.7$
\textbf{DeepTag} & \textbf{48.4} & \textbf{79.9} & \textbf{86.1} & \textbf{62.1} & \textbf{79.8} & \textbf{68.0} & \textbf{82.4} \\ % & $\pm 0.9$
\midrule
& \multicolumn{7}{c}{PP data} \\
\midrule
LSTM & 13.8 & 48.1 & 65.7 & 31.8 & 51.9 & 33.8 & 54.4 \\ 
BLSTM & 13.8 & 47.3 & 66.0 & 35.6 & 57.9 & 36.9 & 58.4 \\ 
DeepTag-M & 17.1 & 53.4 & 68.0 & 37.9 & 59.9 & 40.6 & 61.1 \\ 
\textbf{DeepTag} & \textbf{17.4} & \textbf{56.5} & \textbf{70.3} & \textbf{41.4} & \textbf{62.4} & \textbf{43.2} & \textbf{63.4} \\ 
\bottomrule
\end{tabular}
\begin{flushleft}
Aggregate prediction performance across the 42 categories. BLSTM refers to the multi-task bidirectional LSTM. DeepTag is our best model, and DeepTag-M is the variation with a meta-category loss. EM indicates the exact match ratio, which is the percentage of the clinical notes where the algorithm perfectly predicts all of the disease categories. For example, if a note has three true disease labels, then the algorithm achieves an exact match if it predicts exactly these three labels, no more and no less. For each precision, recall and $F_1$ score, there are two ways to compute an algorithm's performance. First we can take an unweighted average of the score across all the disease categories (unwgt) or we can take an average weighted by the number of test examples in each category (wgt). 
\end{flushleft}
 \label{tab:pp}
\end{table*}
% Macro-F1 is computed wrong...this punishes models that estimate non-0 too harshly. Probably should just expand the coverage...

% ANIE: Subtype category: see how much more does subtype category explain compare to just N.
% ANIE: abbreviation list match; more use of abbreviation
% subtype complexity drives the performance

% \subsubsection*{Correlation with subtype complexity}
% \ashley{it may make sense to move this section to follow the performance paragraphs as this analysis follows the observations of performance in those two datasets}

% for PP, try the difference covered in CSU and see $r^2$ change

% Point out some of them that don't fit the rule
% then maybe analyze 
% \ashley{this is not the most graceful wording, feel free to revise}

\paragraph{More diverse categories are harder to predict.}
After observing the general correlation between number of training examples and per-label $F_1$ scores, we can investigate outliers. These are diseases that have many examples but on which the tagger performed poorly and diseases that have few examples but the tagger performed well. For \emph{disorder of digestive system}, despite having the second highest number of training examples (22,589), both precision and recall are lower than other frequent diseases. We find that this disorder categories covers the second largest number of disease subtypes (694). On the other hand, \emph{disorder of hematopoietic cell proliferation} has the highest $F_1$ score with relatively few training examples ($N = 7294$). This category has only 22 subtypes. Similarly \emph{autoimmune diseases} has  few training examples ($N = 1280$) but it still has a relatively high $F_1$, and it also has only 11 subtypes.

% \ashley{having different body site descriptors does not always cause a change in specific disease code as often these are given by separate modifier, ie. splenic and cutaneous hemangiosarcoma are both hemangiosarcoms with a different body site label.  The major difference is the histologic type, grade, etc.}
The number of subtypes---i.e. the number of different types of lower level codes are mapped to each higher-level disease code---can serve as an indicator for the diversity or specificity of the text descriptions. For a disease like \emph{disorder of digestive system}, it subsumes many different types of diseases such as \emph{periodontal disease}, \emph{hepatic disease}, and \emph{disease of stomach}, which all have
different diagnoses. Similarly, \emph{Neoplasm and/or hamartoma} encapsulates many different histologic types and be categorized as benign, malignant, or unknown, thus resulting in many different lower-level codes (749 codes) being mapped into the same top-level disease code. The tagger needs to associate diverse descriptions to the same high-level label, increasing the difficulty of the tagging.
% TODO: look at this!

% \emph{neoplasm and/or hamartoma}, since it has many histologic types and grades, the tagger needs to associate varied  descriptions to the same high-level label, increasing the difficulty of the tagging.

% about the neoplasm such as benign, malignant, or uncertain all to this meta-disease label, which could be relatively difficult.\ashley{this is a repeat of information above}

% Since we are predicting higher level disease codes, the different types of lower level codes (subtypes) that get mapped to each higher-level disease code is an indicator of how diverse the text descriptions can be. 

% converting codes that are actually applied to the documents into higher-level disease codes, high-level diseases that cover a wider spectrum of lower-level diseases (subtypes) might be harder to tag due to a larger variety of descriptions that the tagger needs to identify.

We hypothesize that disease labels with many subtypes will be difficult for the system to predict. This hypothesis suggests that the number of subtypes within a diagnosis category could explain some of the heterogeneity in DeepTag's performance beyond the heterogeneity due to the training sample size.
%this is a secondary factor that will explain additional variance that is not explained by the primary factor that is the number of training examples.
We conduct a multiple linear regression test with both the number of training examples as well as number of subtypes each label contains as covariates and the $F_1$ score as the outcome. In the regression, the coefficient for number of subtypes is negative with $p < .001$. This indicates that, controlling for the number of training examples, having more subtypes in a disease category makes tagging more challenging and decreases DeepTag's performance on the label.

% \ashley{i would report the number for the coefficient as well as it is a measure of the strength of the effect, basically report the regression equation in full}

\paragraph{Performance on PP}
% investigate these outliers with binning diseases
% maybe binned diseases between CSU and PP are similar on well-performing ones
Next we investigate DeepTag's performance discrepancy between the CSU and PP test data. A primary contributing factor to the discrepancy is that the underlying text in PP is stylistically and functionally different from the text in CSU. Note that DeepTag was only trained on CSU text and was not fine-tuned on PP. The example texts in Figure~\ref{fig:csu-pp-text-examples} illustrate the striking difference. In particular, PP uses many more abbreviations that are not observed in CSU.

%We apply an expert curated dictionary which expands abbreviations on the PP text. %abbreviation dictionary curated by a veterinary expert that expands an abbreviation into fully matched text in CSU. The abbreviation expansion script is triggered 7,475 times.
%After the expansion, 22\% of vocabulary in PP dataset are still not observed in the CSU dataset.  
After filtering out numbers, 15.4\% of words in PP are not found in CSU. Many of the PP specific words appear to be medical acronyms that are not used in CSU or terms that describe test results or medical procedures. Since these vocabulary has no trained and updated word embedding from the CSU dataset, the tagger will not be able to leverage them in the disease tagging process. 
% This could be remedied by generating word embeddings specifically for the veterinarian domain.
% after filtering out common English words, about 70\% of the unmatched words still exist

% (2306 / 10436) \james{explain what these numbers are}
% \james{how do you know they are providing clinically relevant information?}

Despite having many training examples, DeepTag is doing poorly on some very frequent diseases, for example, \emph{neoplasm and/or hamartoma}. 
On the opposite end of the spectrum, the tagger is able to do well for \emph{disorder of auditory system} on both CSU and PP dataset, despite only having a moderate amount of training examples. Besides the main issue of vocabulary mismatch, many subtypes (lower-level codes) that get mapped to a certain disease level code do not exist in PP, and subtypes in PP also might not exist in CSU. We refer to this as the subtype distribution shift.
% 7:
% {'Lipoma (disorder)',
%  'Lymphosarcoma',
%  'Osteosarcoma of bone (disorder)',
%  'Sebaceous adenoma of skin (disorder)',
%  'Secondary malignant neoplasm of liver (disorder)',
%  'Skin papilloma',
%  'Skin tag (disorder)'}
% 11:
% {'Hemangiosarcoma',
%  'Intracranial tumor (disorder)',
%  'Lipoma of trunk',
%  'Lymphoma in remission',
%  'Lymphoma of gastrointestinal tract',
%  'Malignant lymphoma - small lymphocytic (disorder)',
%  'Malignant mast cell neoplasm',
%  'Mast cell tumor',
%  'Meibomian gland adenoma',
%  'Sarcoma',
%  'Small intestine adenocarcinoma'}
% \ashley{did you analyze this? I suspect there will be more benign lesions in the PP dataset, did you look at the 7 and 11 codes specifically}

%\ashley{these must be primarily due to coders using different code levels in the hierarchies for some of these disorders because mast cell tumors, hemangiosarcoma and sarcoma will all show up in the CSU data.  This is, however, a fair observation as there is a difference in how coders code cases, so these things will happen. But I wouldnt emphasize the actual types as it does not appear related to the a true disease distribution discrepancy.  If it comes to it, I have the notes from one CSU tagger and can see how they tag some of these}

For example, In CSU, \emph{neoplasm and/or hamartoma} has 749 observed subtypes. Only 7 out of 749 subtypes are present in PP. Moreover there are 11 subtypes are unique to the PP dataset and are not observed in the CSU training set. These differences appear to be primarily due to differences in how the primary medical codes are applied to the datasets and not significant discrepancies in the types of neoplasias observed.
In addition to the subtype analysis, we note that for rarer diseases, the precision drop between CSU and PP is not as deep as the recall drop. This can be interpreted as the model is fairly confident and precise about the key phrases it discovered from the CSU dataset. However, the PP dataset uses other terms or phrases (that are not covered in the CSU dataset) to describe the disease, resulting in a sharper loss on recall.
% \ashley{you may be asked to provide some examples}
% This is perhaps due to the tagger's ability to capture key phrases that are well-related to the disease.

% There are are few notable exceptions to the general observations of prediction performance in two datasets.  For example, \emph{disorder of hematopoietic cell proliferation} maintains relatively high precision on the PP dataset even though only one subtype is present that was in the CSU dataset.  Also, \emph{disorder of endocrine system} has a lower precision despite having all eight subtypes found in the PP dataset present in CSU.  For these discrepancies, a deeper analysis of the terminologies used to describe the diagnosed diseases may explain the differences observed.

\subsection*{Improvements from disease similarity} % change to similarity???

% \james{This section needs to be reorganized. First, summarize how DeepTag uses hierarchical information and discuss that is better than LSTM and BLSTM. Then say, in addition, we investigated an alternative approach to leverage hierarchy, DeepTag-M, which does... }

% hierarchy
\edit{The 42 SNOMED-CT codes can be naturally grouped into 19 meta-categories; each meta-category corresponds to a subset of diseases that are related to each other\footnote{See Supplementary Material for more information about the grouping.}. For example, the SNOMED-CT codes for ``Disease caused by Arthropod'' and ``Disease caused by Annelida'' belong to the same meta-category, ``Infectious and parasitic diseases''. We designed DeepTag to leverage this hierarchical structure amongst the disease codes. Intuitively, suppose the true disease associated with a note is $A$ and DeepTag mistakenly predicts code $B$. Then its penalty should be larger if $B$ is very different from $A$---i.e., they are in different meta-categories---than if $B$ and $A$ are in the same meta-category. More precisely, we use the grouping of similar codes into meta-categories as a regularization in the training objective of DeepTag. Basic deep learning systems like LSTM and BLSTM do not incorporate this information.}

%DeepTag is designed to leverage the \edit{similarity} between labels.  \footnote{See the supplementary material}.  Based on the \edit{similarity between labels}, we can augment the system with the knowledge that diseases that are under the same parent, which we call a meta-category, should be more similar to each other than diseases that belong to a different parent. 
% The label similarity is an implicit constraint that we can place on the model and it serves as a regularization. Intuitively, Basic deep learning systems like LSTM and BLSTM does not incorporate this information.
% hierarchy

DeepTag uses a $L_2$-based distance objective to place this constraint between disease label embeddings. The objective encourages the embeddings of diseases that are in the same meta-category to be closer to each other than embeddings of diseases across meta-categories. In addition, we investigated another approach that can also leverage similarity: DeepTag-M. This method computes the probability of a parent-level code based on the probability of its children-level codes. Instead of forcing similarity/dissimilarity constraints on disease label embeddings, DeepTag-M encourages the model to make correct prediction on the parent level as well as on the child level. 
% This corresponds to the intuition that we want erroneous predictions to be correct on a broader spectrum

\begin{figure}[!h]
\centering
\includegraphics[scale=0.35]{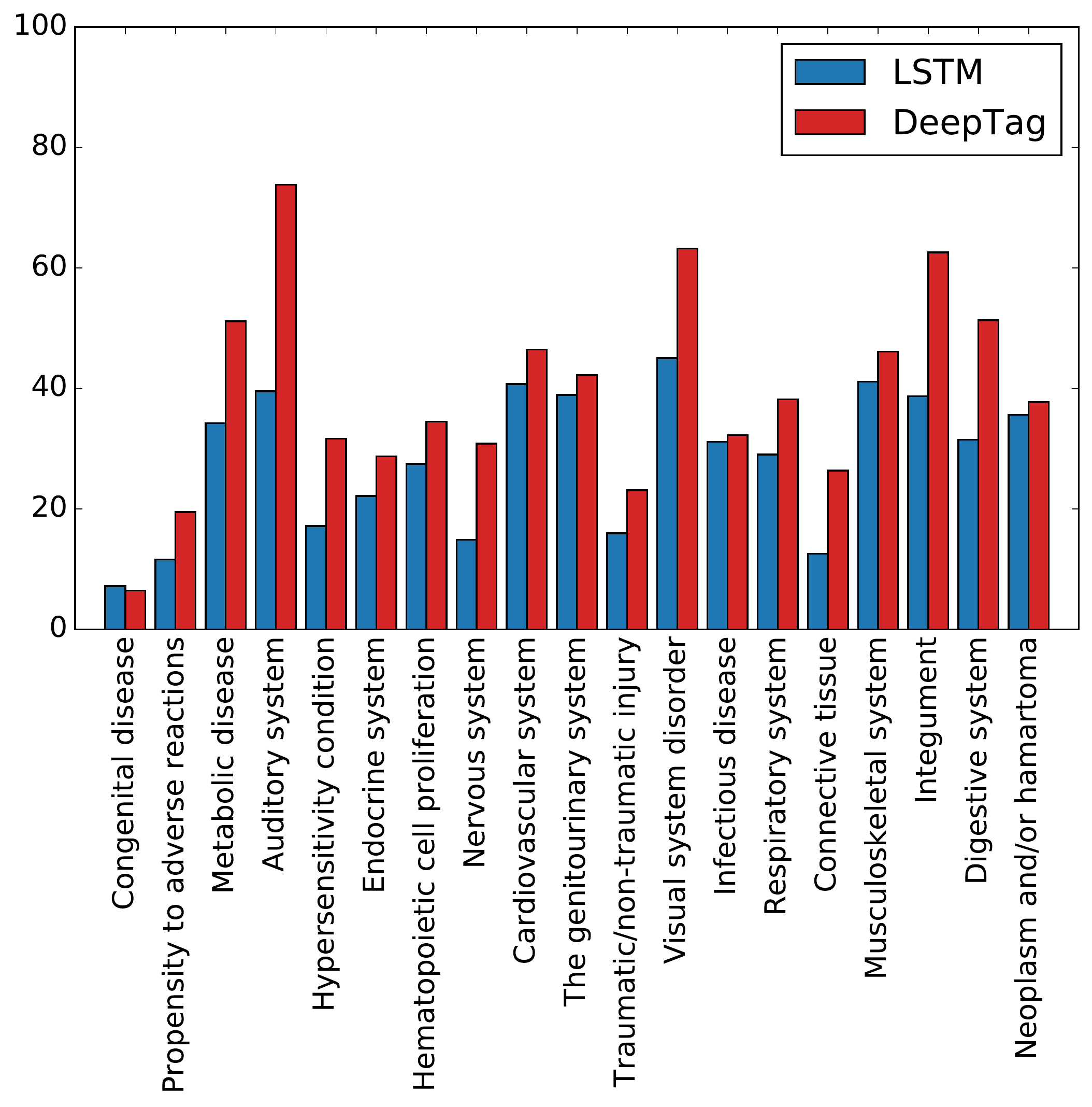}
\caption{{\bf Performance comparison on PP.} 
We compare the per-label $F_1$ score between baseline LSTM model and DeepTag model on the PP dataset. The disease categories are sorted from the least frequent to the most frequent in the training dataset, which comes from CSU.}
\label{fig:pp-comparisons-barplot}
\end{figure}

In Table~\ref{tab:pp}, we compare the performance of DeepTag, DeepTag-M, the standard multi-task LSTM and bidirectional LSTM (BLSTM). 
On the CSU dataset, DeepTag and DeepTag-M perform slightly better (or at the same capacity) compared to the baseline models (LSTM and BLSTM). DeepTag is able to have higher unweighted precision, recall, and $F_1$ score compared to the other models, indicating its ability to have good performance on a wide spectrum of diseases. The importance of leveraging similarity is shown on the PP dataset (Figure~\ref{fig:pp-comparisons-barplot}).  Since it is out-of-domain, expert defined disease similarity provide much-needed regularization to make both DeepTag and DeepTag-M outperform baseline models by a substantial margin, with DeepTag being the overall best model. 

% From the Table~\ref{tab:pp}, we can see that augmenting the algorithm to leverage hierarchy between diseases results performs quite similarly on the original dataset. We do observe that the model with meta-label prediction loss (\textbf{+M}) performs quite well in the lower regime. However, the major improvement we observe is between LSTM and bidirectional LSTM (BLSTM) on the CSU data. In order to properly evaluate the tagger's performance, we compute both weighted and unweighted $F_1$ scores. First, we compute a weighted average across all disease labels weighted by the label frequency (more frequent disease gets higher weight).  Second, we compute a simple average score over all diseases that do not have a zero score. \james{Use weighted and unweighted instead of micro and macro. Explain EM.}

% Hierarchy-related penalties encourage model to either keep parameters in close distance for labels that are similar (cluster penalty), or create ``virtual'' signals for rarer labels even when they are not present in the example (meta label prediction loss \textbf{+M}). These strategies could very well obfuscate the decision boundary that the model was supposed to learn. Bidirectional LSTM with cluster penalty (DeepTag) does not outperform the bidirectional LSTM (BLSTM) on many labels.

% \ashley{the x-axis labels for the Figure 3 plot have inconsistent capitalization}

\subsection*{Learning to abstain}

% calibration plot
% \james{As I said before, this is too much detail for here and should be moved to Methods or Appendix. Summarize at a high level what learning to abstain is trying to do here. }

Augmenting a tagging system with the ability to abstain (decline to assign codes) can foster human-machine collaboration. When the system does not have enough confidence to make decisions, it has the option to defer to its human counterparts. This aspect is important in DeepTag because after tagging the documents, further analysis from various parties might be conducted on the tagged documents such as investigating the prevalence of certain specific diseases. In order to not mislead further clinical research, having the ability to abstain from making very erroneous predictions and ensuring highly precise tagging is an important feature.

We developed an additional abstention wrapper on top of DeepTag that we call DeepTag-abstain. The module learns to estimate how well the DeepTag system will perform on a document based on the predicted categories DeepTag makes on the document as well as DeepTag's internal confidence on the predictions. 

We compare DeepTag-abstain to an intuitive baseline where an abstention score is simply computed by the confidence associated with the diagnosis code assignments. In order to evaluate how well DeepTag-abstain performs compared to the baseline, we compute an abstention priority score for each document. A document with higher abstention priority score will be removed earlier than a document with low score. We then compute the weighted average of $F_1$ and exact match ratio (EM) for all the documents that are not removed.

For both baseline and DeepTag-abstain, we specify a proportion of the documents need to be removed. We adjust the dropped portion from 0 to 0.9 (dropping 90\% of the examples at the high end). An abstention method that can drop more erroneously tagged documents earlier will observe a faster increase in its performance, corresponding to a curve with steeper slope. 
% \ashley{what is the main point you are trying to make here?}

% Figure~\ref{fig:abs-best-curve} shows the improvement curve of exact match ratio (EM) and micro-$F_1$ (F1). 

% The dotted line indicates the confidence-based abstention which we computed on the test set, which we used as the baseline and is constant among four sub-plots. We adjusted the dropped proportion from 0 to 0.9 (dropping 90\% of the test examples), and observed the improvement of exact match ratio and micro-averaged $F_1$ score. Systems with better abstention priority scores will be able to abstain ``worse'' examples, therefore gain steeper increase in both metrics.

\begin{figure}[!h]
    \centering
    \includegraphics[scale=0.4]{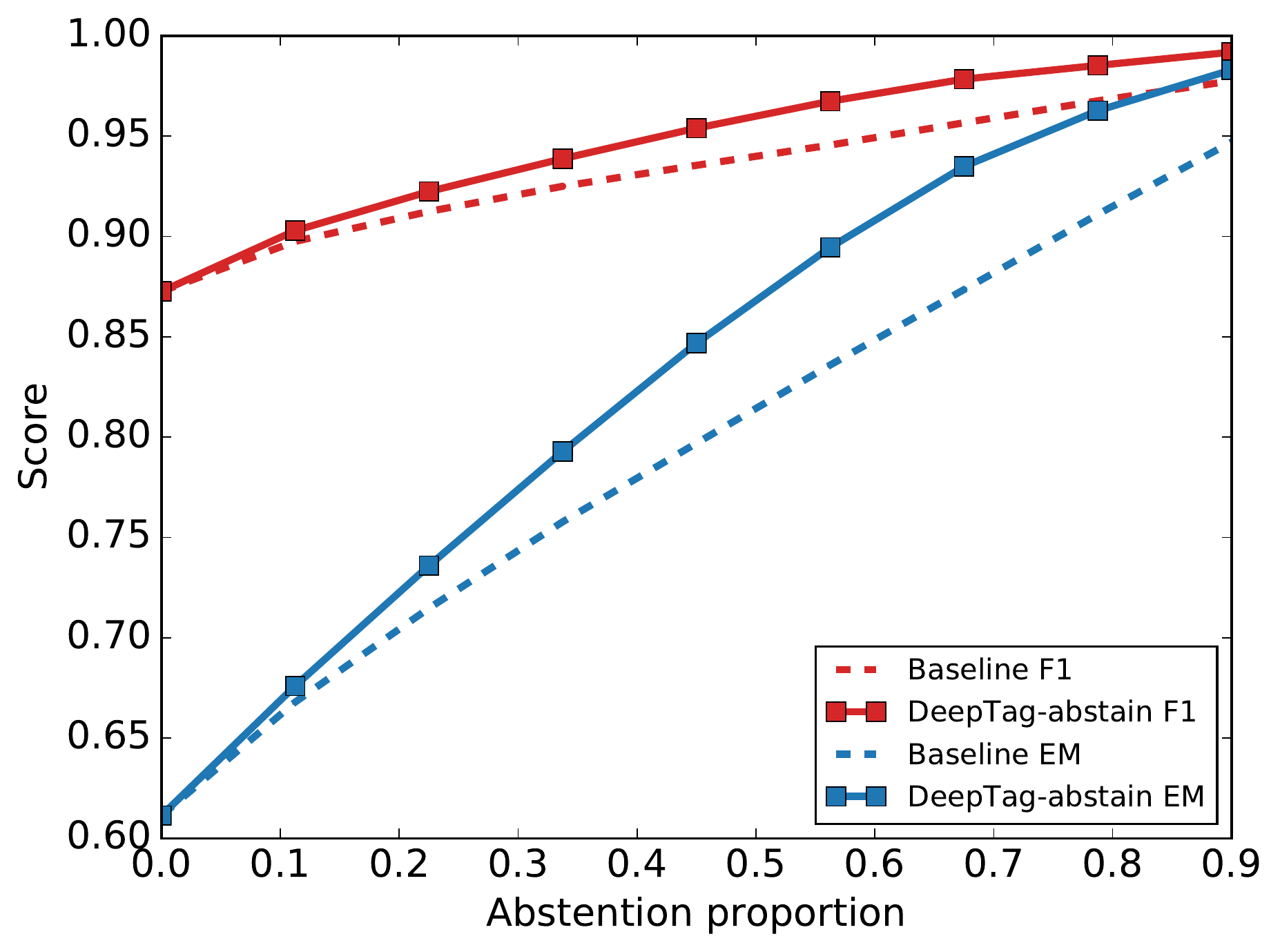}
    %~ %add desired spacing between images, e. g. ~, \quad, \qquad, \hfill etc. 
\caption{{\bf Comparison of the abstention models. } DeepTag-abstain is the abstention priority score estimator that uses confidence scores as input and estimate instance-level accuracy of a given document. Baseline refers to the abstention scheme where the instance-level abstention priority score is computed from individual label confidence scores without any learning. As a greater proportion of the examples are abstained from, the performance---$F_1$ and Exact Match (EM)---of both methods improve. DeepTag-abstain shows faster improvement, indicating that it learns to abstain in more difficult cases.}
\label{fig:abs-best-curve}
\end{figure}

DeepTag-abstain demonstrates a substantial improvement over the baseline in Figure~\ref{fig:abs-best-curve}. The baseline here is the natural approach that abstains based on the original DeepTag's uncertainty at the last layer. 
 DeepTag-abstain is a more powerful approach that \emph{learns} where to abstain based on the model's internal representation of the input text. 
We note that not all learning to abstain schemes are able to out-perform the baseline. The details of module design and improvement curve for the rest of the modules can be seen in Appendix Fig~\ref{fig:abs-improv-curve}.  

%Noticeably, taking confidence scores as input and estimating accuracy gives the best result, outperforming other models as well as the baseline.

% \begin{figure}[!h]
% % width=\textwidth
% \centering
% \includegraphics[scale=0.4]{images/best-abstainsion.png}
% \caption{{\bf Comparison of learned abstention model.} Compare the best learn to abstain models. ``LTA Conf Accu'' is the abstention priority score estimator that uses confidence scores as input and estimate instance-level accuracy of a given document. $r^2$ of this model is 0.2537.  ``LTA Pred Loss'' takes  probability of each label as input and predict the average binary cross entropy loss of the document. $r^2$ of this model is 0.2415.
% }
% \label{fig:abs-best-curve}
% \end{figure}

% \james{reorder the legend so that all the orange legend is on top together and the blues ones at the bottom.}
% We conduct a linear regression analysis of the model's predicted output and true output. $r^2$ of this model is 0.25.
% \james{What is this $r^2$ referring to? }
% \james{Change the x-label to abstention proportion.}
\section*{Discussion}

In this study, we developed a multi-label classification algorithm for veterinary clinical text, which \edit{represents a medical domain with an under-resourced medical coding infrastructure}. In order to improve the performance of DeepTag on diseases with rare occurrences, we investigated with loss augmentation strategies that leverage the hierarchical structure of the disease categories. These augmentations provide gains over the LSTM and BLSTM baselines, which are common methods used for these types of prediction tasks. We also experimented with different methodologies to allow the model to learn to abstain on examples where the model is not confident in the predictions. We demonstrate that learned abstention rules outperform manually set rules.

%Add ashley comments;
Our work demonstrates novel methods for applying broad disease category labels to clinical records as well as applying those trained algorithms to an external dataset in order to examine cross-hospital generalization. We also demonstrate means to allow human domain experts to use their judgement where automated taggers have a high level of uncertainty in order to improve the overall workflow. We confirm that cross-hospital generalization is a significant concern for learned tagging systems to be deployed in real world implementations that may vary substantially from the data on which they were trained. Even though our work attempts to mitigate this problem, there is significant research to be done in optimizing methods for domain adaptation.
Our current work is important not only for veterinary medical records, which are rarely coded, but also may have implications for human medical records in countries with limited coding infrastructure and which are important regions of the world for public health surveillance and clinical research. 

%This paper presents DeepTag, an algorithm that infers SNOMED disease codes from veterinary notes. DeepTag leverages hierarchical structures among the diseases to improve accuracy. It also uses an learning approach to abstain from notes that it is unsure about. 
%Our major goal in this work was to evaluate methodological advances to improve assigning of broad disease labels to free text by utilizing intrinsic terminology hierarchies and identify ways to allow for incorporation of human domain expertise when it is needed to improve labeling predictions.  

%\paragraph{Generalization evaluation} 

There are several aspects of the data that may have limited our ability to apply methods from our training set to our external validation set. Private veterinary practices often have data records that closely resemble the PP dataset used to evaluate our methods here. However, the large annotated dataset we used for training is from an academic institution (as these are, largely, the institutions that have dedicated medical coding staff). As can be seen from Table~\ref{tab:pp}, the performance drop due to domain mismatch is non-negligible. The domain shift comes from two parts. First, text style mismatch -- private commercial notes use more abbreviations and tend to include many procedural examinations (even though many are non-informative or non-diagnostic). This requires the model to learn beyond keyword or phrase matching. Second, label distribution mismatch -- the CSU training dataset focuses largely on neoplasm and several other tumor-related diseases, largely due to the fact that the CSU hospital is a regional tertiary referral center for cancer and cancer represents nearly 30\% of the caseload. Other practices will have datasets composed of labels that appear with different frequencies, depending on the specializations of that particular practice. A very important path forward is to use learning algorithms that are robust to domain shift, and experimenting with unsupervised representation learning to mitigate the domain shift between academic datasets and private practice datasets.

 %Even though confidence scores are transformed from label probability via a deterministic piece-wise function, a three-layer neural network cannot approximate this function well with the number of examples in the training set. 

% Interestingly, the best performing model for abstention takes confidence scores as input and it outperforms using the direct label probability as input.  Another noticeable finding is that even though label probabilities are obtained after applying sigmoid function over the logit, it is still worse than label probabilities. The input that completely under-performs the baseline is the hidden representation meaning that without change in loss, the transformed input itself does not contain enough information about the final loss or accuracy without the binary classifier parameters.

%\paragraph{Choice of prediction labels} 

% A possible extension of the research is to learn to also tag specific important procedures, drug reactions, etc. from the same set of notes.  Additionally,
Currently we are predicting top-level SNOMED-CT disease codes, which are not the SNOMED-CT codes that have been directly annotated on the dataset. Many of the SNOMED-CT codes that are applied to clinical records are categorized as 'Findings' that are not actual 'Disorders' as the actual diagnosis of a patient may not be clear at the time the codes are applied.  One example is an animal that is evaluated for 'vomiting' and the actual cause is not determined, may have a code of 'vomiting(finding)'(300359004) applied and not 'vomiting(disorder)'(422400008) and these 'non-disorder' disease codes are not evaluated in our current work.  However, these are an important subset of codes and represent another means to identify particular patient cohorts with particular clinical signs or presentations, vs. diagnosed disorders. 

\edit{Another future direction for the abstention branch of this work is to factor human cost and annotation accuracy into the model and only defer when the model believes that human experts will bring improvement to the result within an acceptable amount of cost. This is an interesting direction for experimentation.}

% \edit{Another future direction for the abstention branch of this work is to factor human cost and annotation accuracy into the model and only defer when the model believes that human experts will bring improvement to the result within an acceptable amount of cost. This is out of scope for our work, and we point readers to Werling et al. for human-machine joint optimization in a non-clinical domain \cite{werling2015job}.}

%other future directions or ways to improve the algorithms?

% In NPJ Digital Meidicine: the last paragraphs of discussion IS the conclusion section!!

% rewrite this?

\section*{Methods}

% Table 1: describe how many records are for different species (cat, dog)
% Sex, age, clinical descriptors (Google paper is like this too)
% demographics, a clinical table as table 1!

\subsection*{Datasets}

\paragraph{Colorado State University dataset}
The CSU dataset contains discharge summaries as well as applied diagnostic codes for clinical patients from the Colorado State University College of Veterinary Medicine and Biomedical Sciences. 
This institution is a tertiary referral center with an active and nationally recognized cancer center. Because of this, the CSU dataset over-represents cancer-related diseases. Rare disease categories in the CSU dataset are diseases like pregnancy, perinatal and mental disorders, but these are also rare in the larger veterinary population as a whole and do not represent a dataset bias. Overall, there are 112,558 unique discharge summaries in CSU dataset.  We split this dataset into training, validation, and test set by 0.9/0.05/0.05. 

\paragraph{Private Practice dataset}
An external validation dataset was obtained from a regional private practice (PP).  These records did not have diagnostic codes available and only approximately 3\% of these records had free text diagnoses applied by the attending clinician. Two veterinary domain experts applied SNOMED-CT disease codes to a subset of these records and achieved consensus on the records used for validation. This dataset (PP) is used for external validation of algorithms developed using the CSU dataset. There are 586 documents in this external validation dataset.
% \ashley{I'm checking with Devin on the 586 vs. 688 records issue}

\subsection*{Data processing}

% We preprocessed the text data with standard English tokenization from SpaCy~\footnote{\url{https://spacy.io/}}. We used regex pattern matching to remove HTML entities, additional white spaces, and URL links. 

% Application of disease codes (Ashley)
Documents in our corpus have been tagged with SNOMED-CT codes that describe the clinical conditions present at the time of the visit being annotated.  Annotations are applied from the SNOMED-CT veterinary extension (SNOMEDCT\_VET), which is fully compatible and is an extension of the International SNOMED-CT edition.  It can be accessed in a dedicated browser and is maintained by the Veterinary Terminology Services Laboratory at the Virginia-Maryland Regional College of Veterinary Medicine\footnote{\url{http://vtsl.vetmed.vt.edu/default.cfm}}.  Medical coders applying diagnostic codes are either veterinarians or trained medical coders with expertise in the veterinary domain and the SNOMED terminology. \edit{The medical coding staff at CSU utilize post-coordinated expressions, where required, for annotations to fully describe a diagnosed condition. For this work, we only considered the core disease codes and not the subsequent modifiers for training our models.  The PP dataset was similarly coded using post-coordinated terms following consultation with coding staff at multiple academic institutions that annotate records using SNOMED-CT. We further grouped the 42 SNOMED-CT codes into 19 meta-categories. More details of this grouping are provided in the Supplement.}

\subsection*{Difference in data structures}
Due to the inherent differences in clinical notes/discharge summaries prepared for patients in an academic setting compared to the shorter 'SOAP' format notes (Subjective, Objective, Assessment, Plan) prepared in private practice, there are substantial differences in the format as well as the writing style and level of detail between these two datasets. In addition, the private practice records exhibit significant differences in record styles between clinicians, as some clinicians use standardized forms while others use abbreviated clinical notes containing only references to abnormal clinical findings. 

As can be seen in Fig~\ref{fig:doc-lengths-label-dist}, both dataset have more than 80\% documents associated with more than one label, and in terms of document length distribution, PP dataset document is much shorter than CSU dataset, while the average PP document length is 191 words. The average CSU document length is 325 words.

% \james{Make the ticks line up with the corresponding bars. Also, double check the y values of the left plot; seems too low?} \ashley{suggested x-axis labels: 'length of document(words)' and 'number of disease labels per document'}

In order to bridge the gap between the two domains, we additionally use a curated veterinarian abbreviation list that maps an abbreviation to its full text. We include this abbreviation list in our supplementary materials. 

% \ashley{should this be 'supplementary material?}

\subsection*{Algorithm development and analysis}

We trained our modeling algorithm on CSU dataset and evaluated on a held-out portion of data from the CSU dataset as well as the PP dataset. 
% CSU and PP dataset vary significantly in terms of style and length of document.
We formulated our base model to be a recurrent neural network with long short-term memory cells (LSTMs). We additionally decided to run this recurrent neural network on both the forward direction and backward direction of the document (bidrectional), as is found beneficial in Graves et al. \cite{graves2005bidirectional}. \edit{We then built 42 independent binary classifiers to predict the existence of each label. This is the architecture found most useful in multi-label classification literature \cite{kavuluru2015empirical}. The model is trained jointly with binary cross entropy loss.} We then augmented this baseline model with two losses: cluster penalty \cite{jacob2009clustered} and a novel meta-label prediction loss to leverage human expert knowledge in how semantically related these disease labels are. 
% \edit{We chose our architecture primarily based on the performance on the CSU dataset. We ran a convolutional neural network (CNN) as another baseline. Details of these models are reported in the supplementary material.}

We tuned the clustering penalty hyperparameters $\gamma_{\mathrm{norm}}$, $\gamma_{\mathrm{within}}$ and $\gamma_{\mathrm{between}}$, and our search range was [1e-1, 1e-5]. We also tuned the meta-label prediction loss hyperparameter $\beta$ in a similar range.

% As macro-averaged scores are likely to be influenced by outliers, we compute the average of non-zero metrics, ignoring tagger's performance when it is completely false. Details of our model and loss augmentation can be found in Supplement.

\section*{Data availability}
The data that support the findings of this study are available from Colorado State University College of Veterinary Medicine and a private practice veterinary hospital near San Francisco, but restrictions apply to the availability of these data,
which were made available to Stanford for the current study, and so are not publicly available. Data are however available from the authors upon
reasonable request and with permission of Colorado State University College of Veterinary Medicine and the private hospital.

\section*{Code availability}
DeepTag is freely available at \url{https://github.com/windweller/DeepTag}.

\section*{Acknowledgements}

We would like to acknowledge the help of Devin Johnsen for her help in annotating the private practice records used in this work. We also want to thank Selina Dwight and Matthew Wright for helpful feedback. 
Our work is funded by the Chan-Zuckerberg Investigator Program and  National Science Foundation (NSF) Grant CRII 1657155.
% Grant or contribution numbers may be acknowledged.

\section*{Author contributions statement}

% Must include all authors, identified by initials, for example:
A.N. performed the natural language processing work, designed and built DeepTag, and performed all of the experiments.  A.Z. acquired the data, established the meta-hierarchies, organized and aided in annotation of the private practice data, and provided feedback on the NLP and machine learning outputs.  R.P. provided access to the training data. A.P. aided in initial data acquisition and provided feedback on the project focus and strategy. M.R., S.D. and C.B. provided feedback on the project. J.Z. designed and supervised the project. A.N., A.Z. and J.Z. wrote the paper.

\section*{Additional information}
\textbf{Competing Interests}: the authors declare no competing interests.
 
%To include, in this order: \textbf{Accession codes} (where applicable); \textbf{Competing financial interests} (mandatory statement). 

%The corresponding author is responsible for submitting a \href{http://www.nature.com/srep/policies/index.html#competing}{competing financial interests statement} on behalf of all authors of the paper. This statement must be included in the submitted article file.

\bibliography{sample}

% \noindent LaTeX formats citations and references automatically using the bibliography records in your .bib file, which you can edit via the project menu. Use the cite command for an inline citation, e.g.  \cite{Figueredo:2009dg}. 

% \begin{figure}[ht]
% \centering
% \includegraphics[width=\linewidth]{stream}
% \caption{Legend (350 words max). Example legend text.}
% \label{fig:stream}
% \end{figure}

% \begin{table}[ht]
% \centering
% \begin{tabular}{|l|l|l|}
% \hline
% Condition & n & p \\
% \hline
% A & 5 & 0.1 \\
% \hline
% B & 10 & 0.01 \\
% \hline
% \end{tabular}
% \caption{\label{tab:example}Legend (350 words max). Example legend text.}
% \end{table}

% Figures and tables can be referenced in LaTeX using the ref command, e.g. Figure \ref{fig:stream} and Table \ref{tab:example}.

\clearpage

\appendix

\setcounter{figure}{0}
\renewcommand{\thefigure}{S\arabic{figure}}
\setcounter{table}{0}  % for future table handling
\renewcommand{\thetable}{S\arabic{table}}
\section*{Supplementary Materials}

%\subsection*{Development of SNOMED-CT to ICD9 correspondence}

%In order to link top-level SNOMED-CT disease codes into related categories for the purposes of the learning with label similarity for this work, we have arranged a subset (n=41) of the 95 SNOMED disease codes (direct children of conceptID 64572001) into a hierarchy that mimics that of top level ICD9 codes.  Excluded codes are those that are not directly relevant to a veterinary classification task due to the infrequency of certain disorders (ex: 278919001: \emph{Communication disorders} or ex: 242028000: \emph{Weightlessness}) or code categories are functionally redundant due to the structure of the hierarchy (i.e., diseases categorized under \emph{diabetic complication} also map to the relevant body system categories, thus 'diabetic cataracts' are captured under \emph{visual system disorders}). This mapping of high level disease hierarchies allows veterinary diseases to be directly compared to human medical notes that are more commonly coded in ICD9 or ICD10 than using SNOMED-CT. Currently available SNOMED-CT to ICD9 maps do not include the SNOMEDCT-VET extension and also do not map the higher level disease codes used in this work. The mapping of top level SNOMED disease codes to the relevant ICD9 codes is provided as a supplemental table. Mappings were reviewed by experienced veterinary medical coders familiar with the SNOMED-CT hierarchy.

\subsection*{CSU discharge summary format}

\edit{The Colorado State University discharge summaries contain multiple data fields, including: History, Assessment, Diagnosis, Prognosis, FollowUpPlan, ProceduresAndTreatments, PendingDiagnostics, PendingDiagnosticsComments, Diet, Exercise, DischargeStatus, DischargeDate, Medications, AdditionalInstructions DrugWithdrawal, RecheckVisits, Complications, MedicalComplications, SurgicalComplications and AnesthesiaComplications. We filtered out fields with many null entries as well as the diagnosis related fields, since this is not present in the private practice data. The remaining fields---History, Assessment, Prognosis, DischargeStatus and Medications---were used as the input to train the models.}

\subsection*{Model description}

We formulate the problem of veterinary disease tagging as a multi-label classification problem. Given a veterinary record $\capital{X}$, which contains detailed description of the diagnosis, we try to infer a subset of diseases $\bm{y} \in \mathcal{\bm{Y}}$, given a pre-defined set of diseases $\mathcal{\bm{Y}}$. The problem of inferring a subset of labels can be viewed as a series of independent binary prediction problems~\cite{sorower2010literature}. The binary classifier learns to predict whether a tag $y_i$ exists or not for $i = 1,...,m$, where $m = |\mathcal{\bm{Y}}|$.

Our learning system has two components: a text processing module and tag prediction module. Our text processing module will use long-short-term memory networks (LSTMs) which have demonstrated their effectiveness in learning implicit language patterns from the text \cite{mikolov2012statistical}. Our tag prediction module will consist of binary classifiers that are parameterized independently. 

A long-short-term memory networks is a recurrent neural network with LSTM cell. It takes one word as input, as well as the previous cell and hidden state. Given a sequence of word embeddings $x_1, ..., x_T$, the recurrent computation of LSTM networks at a time step $t$ can be described in Eq~\ref{eq:lstm}, where $\sigma$ is the sigmoid function $\sigma = 1 / (1 + e^{-x})$, and $\tanh$ is the hyperbolic tangent function. We use $\odot$ to indicate the hadamard product.

\begin{equation}
\begin{split}
f_t & = \sigma(W_f x_t + V_f h_{t - 1} + b_f) \\
i_t & = \sigma(W_i x_t + V_i h_{t - 1} + b_i) \\
o_t & = \sigma(W_o x_t + V_o h_{t - 1} + b_o) \\
\tilde{c}_t & = \tanh(W_c x_t + V_c h_{t - 1} + b_c) \\
c_t & = f_t \odot c_{t - 1} + i_t \odot \tilde{c}_t \\
h_t & = o_t \odot \tanh(c_t)
\end{split}
\label{eq:lstm}
\end{equation}

% We simplify this set of equations to $h_t = \text{LSTM}(x_t | \theta)$ where $\theta$ is all the weight and bias parameters in the computation.

An extension of this recurrent neural network with LSTM cell is to introduce bidirectional passes \cite{graves2005bidirectional}. Graves et al. shows that introducing bidrectional passes, it can effectively eliminate problems such as retaining long-term dependency when the document is very long. We parameterize two LSTM cells with different set of parameters, one cell is used in forward pass where the sequence is passed in sequentially from the beginning $\{x_1, ..., x_T\}$ , one cell is used for backward pass, where the sequence is passed in with reversed ordering $\{x_T, ..., x_1\}$. At the end of both passes, bidirectional LSTM will output two hidden states represents each input $x_t$, and we stack these two hidden states as our new hidden state for this input $ h_t = [\overrightarrow{h_t};\overleftarrow{h_t}]$.

After computing hidden states over the entire documents, we introduce global max pooling over the hidden states, as suggested by Collobert \& Weston \cite{collobert2008unified} so that the hidden states will aggregate information from the entire documents. Assuming the dimension of hidden state is $d$, global max pooling apply an element-wise maximum operation over the temporal dimension of the hidden state matrix, described in Eq~\ref{eq:gmp}.

\begin{equation}
\begin{aligned}
H &= [h_1, ..., h_T] \text{, } H \in \reals^{T \times d} \\
c_j &= \max(H_{j}) \text{, for } j = 1, ..., d\\
\end{aligned}
\label{eq:gmp}
\end{equation}

% binary classifiers
Then we define a binary classifier for each label in our pre-defined set. The binary classifier takes in a vector $\bm{c}$ that represents the veterinary record and outputs a sufficient statistic for the Bernoulli probability distribution indicating the probability of whether a tag should is predicted. For $i = 1, ..., m$:

\begin{equation}
\begin{aligned}
% g_i(\bm{c}) = \hat{y}_i = \sigma(\theta_i^\intercal \bm{c})\\
p(y_i) = \hat{y}_i = \sigma(\theta_i^\intercal \bm{c})\\
\end{aligned}
\label{eq:binary}
\end{equation}

% loss
We use binary cross entropy loss averaged across all labels as the training loss. Given the binary predictions from the model $\bm{\hat y} \in [0, 1]^m$ and correct one-hot label $\bm{y} \in \{0, 1\}^m$, binary cross entropy loss is written as follow:

\begin{equation}
\begin{aligned}
\mathcal{L}_{\text{BCE}}(\bm{\hat{y}}, \bm{y}) = - \frac{1}{m} \sum_{i=1}^m y_i \log (\hat y_i) + (1 - y_i)\log(1- \hat y_i) \\
\end{aligned}
\end{equation}

As usual, the decision boundary in our model is 0.5. We can generate a list of predicted label by applying a decision function $d$:

\begin{equation}
\begin{aligned}
d(\hat y_i) =
\begin{cases} 
      0 & \text{if } \hat y_i \leq 0.5 \\
      1 & \text{if } \hat y_i > 0.5 \\
   \end{cases}
\end{aligned}
\end{equation}

% \edit{As an alternative baseline, we trained a convolutional neural network (CNN) for document classification. We formulated this network to be similar to the model used by Kim \cite{kim2014convolutional} and Baumel et al \cite{baumel2017multi}, which is a single layer of convolution followed by ReLU activation function as well as global max pooling in the end. We set the filter size to produce sentence vector the same dimension as a bidirectional LSTM.}

\subsection*{Leveraging disease similarity}

% talk about clustering penalty
% and meta-label penalties here

% how we created meta-labels by maping SNOMED disease to ICD-9 codes (ASHELY) (K meta-clusters) - discussed above

We introduce two penalties that are inspired by the implicit relationships between the SNOMED-CT disease codes that we refer to as meta-labels or clusters. By augmenting our loss with these two penalties, we aim to increase model's ability to predict labels that have fewer instances. In the result section, we refer to model trained with cluster penalty as ``DeepTag'', and model trained with meta-label prediction loss as ``DeepTag-M''.

\paragraph{Cluster penalty} 

After defining the meta-labels for the SNOMED-CT disease tags, we can use techniques the from multi-task learning literature. Jacob et al.\cite{jacob2009clustered}
proposed a hypothesis that if two tasks are similar, the task-specific parameters for these two tasks should be close in vector space, vice versa.

Following Jacob et. al, we can first compute the mean vector of all tasks $\bar \theta =  \frac{1}{m}\sum_{i=1}^m \theta_i$. We can define $\mathcal{J}(k) \subset \{1, ..., m\}$, where $\mathcal{J}(k)$ is a set of labels that belong to cluster $k$. Then we can compute the mean vector for each cluster of tasks: for $k=1,...,K$, $\bar \theta_k = \frac{1}{|\mathcal{J}(k)|} \sum_{i=1}^{|\mathcal{J}(k)|} \theta_i$.

The within-cluster closeness constraint $\Omega_\text{within}$ can be computed as the distance between task specific weight vectors and the cluster mean vector $\bar \theta_k$. $\Omega_\text{between}$ can be computed as the distance between $\bar \theta_k$ and $\bar\theta$.
We formulate this as an additional loss term $\Omega(\Theta)$, and allow three hyperparameter $\gamma_{\mathrm{norm}}$, $\gamma_{\mathrm{within}}$ and $\gamma_{\mathrm{between}}$ to control the strength of this penalty.

% When the similarity function is dot product, we have the problem that the similarity is unbounded. To mitigate this issue, we can further introduce a $L_2$ norm penalty over all task parameters. We further define $\mathcal{J}(k) \subset \{1, ..., m\}$, where $\mathcal{J}(k)$ is a set of labels that belong to meta cluster $k$. 

\begin{equation}
\begin{aligned}
\Omega_{\mathrm{norm}} &= \sum_{i=1}^m ||\theta_i||^2 \\
\Omega_{\mathrm{between}} &= \sum_{k=1}^K ||\bar\theta_k - \bar\theta||^2  \\
\Omega_{\mathrm{within}} &= \sum_{k=1}^K \sum_{i \in \mathcal{J}(k)} ||\theta_i - \bar \theta_k||^2 \\
% \Omega_{\mathrm{between}} &= \sum_{k=1}^K \sum_{\substack{i=1 \\ i \not\in \mathcal{J}(k)}}^m S(\theta_i, \tilde \theta_k) \\
\end{aligned}
\label{eq:clustering}
\end{equation}

\paragraph{Meta-label prediction loss} 

We propose an additional penalty following the intuition that we want the model to make accurate predictions for the broad category even though mistakes can be made on the fine-grained level. Meta labels are created by examining whether any disease label under this meta label has been marked as tagged. Following the same logic,
since the disease labels are predicted independently, we can compute the probability of the presence of a meta label $\tilde y_k$ from the probability of disease labels that belong to this meta label.
% Eq~\ref{eq:meta}. 

\begin{equation}
\begin{aligned}
p(\tilde{y}_k) &= 1 - \prod_{i \in \mathcal{J}(k)} (1 - p(y_i))  \\
&= 1 - \prod_{i \in \mathcal{J}(k)} (1 - \sigma(\theta_i^\intercal \bm{c})) \\
\end{aligned}
\label{eq:meta}
\end{equation}

After computing the probability of presence of each meta-label, given the set of meta labels $\bm{\tilde y}$ that are created from our true set of labels $\bm{y}$, we can then compute the binary cross entropy loss between the model's estimation on meta label probability and true meta labels in Eq~\ref{eq:meta-loss}. We use $\beta$ to adjust the strength of this penalty.

\begin{equation}
\begin{aligned}
\mathcal{L}_{\text{meta}}(p(\bm{\tilde{y}}), \bm{\tilde y}) = - \frac{1}{K} \sum_{k=1}^K \tilde{y}_k \log (p(\tilde{y}_k)) \\ +  (1 - \tilde{y_k})\log(1- p(\tilde{y}_k)) \\
\end{aligned}
\label{eq:meta-loss}
\end{equation}

\subsection*{Learning to abstain}

% distance to 0.5
% convert to example-based score
% learn to abstain on example-level using loss / predicting loss (show if this is better)
% need to say example-based meaning that we are dropping all labels from an example, can't abstain normally.

% maybe add some kind of introduction...

In practice, it is often desirable for the model to forfeit the prediction if the prediction is likely to be incorrect. When the method is used in collaboration with human experts, the model can just defer difficult cases to them, fostering human-computer collaboration. However, this is still an under-explored field in machine learning, and previous research has focused largely on binary-class single-label classification \cite{cortes2016learning}.
We formally describe the set-up and our learning-based approach in the following sections, and extend relevant discussion to a multi-label setting.

We propose two abstention settings. Each setting will compute a score $\alpha$ for each document, which we refer to as the abstention priority score. We can then rank these documents using this score $\alpha$.
When user specifies a percentage of documents to be dropped, 
documents that have high $\alpha$ will be dropped first.
% In order to decide what examples should be abstained and what examples should be kept, we need to compute a score $\alpha$ for each example, and we refer to $\alpha$ as the abstention priority score. 

\paragraph{Confidence-based abstention}
% need to show calibration plot...

Since our model already outputs a probability for each label, if our model is well-calibrated, meaning that the output probability satisfies the following constraint in Eq~\ref{eq:calibration}, then our probability should reflect how uncertain the model is about the output.

\begin{equation}
\begin{aligned}
\text{P}_{x, y \sim \mathcal{D}}[y=1 | f_t(x) = p] = p \text{ } \forall p \in [0, 1] \text{ and } \forall t \\
\end{aligned}
\label{eq:calibration}
\end{equation}

The notion of calibration means that when the model thinks the chance of a given prediction to be correct is p\%, we collect all instances that the model gives such probability, and the model in total will be correct p\% of the time. A well-calibrated model's output probability corresponds to the model's confidence/certainty on how correct its prediction is. Previous research has shown that binary classifiers with sigmoid scoring function and cross-entropy loss are often well-calibrated \cite{niculescu2005predicting}.

Given calibrated $\{p(y_1), ..., p(y_m)\}$, we want to compute how confident the model is on these predictions. Noticeably,  For each prediction, the model is more confident if $p(y_i)$ is farther away from 0.5. Based on this observation, we can convert the probability into a confidence score with function $g$:

\begin{equation}
\begin{aligned}
g(p(y_i)) =
\begin{cases} 
      1 - p(y_i) & \text{if } p(y_i) \leq 0.5 \\
      p(y_i) & \text{if } p(y_i) > 0.5 \\
   \end{cases}
\end{aligned}
\end{equation}

We can now compute the probability of the model getting $k$ labels correct on a single example. We choose all subsets from the entire label set, and compute the probability of a chosen subset to be correct as well as the probability of the not chosen $(m-k)$ labels to be incorrect.

\begin{equation}
\begin{aligned}
\alpha_{\text{conf}} = \sum_{\substack{ I \subset \{1, ..., m\} \\ |I| = k}} (\prod_{i \in I} g(p(y_i))) (\prod_{j \not\in I} 1 - g(p(y_j))) \\
\end{aligned}
\label{eq:abs-baseline}
\end{equation}

\begin{figure*}[!h]
\centering
\includegraphics[scale=0.35]{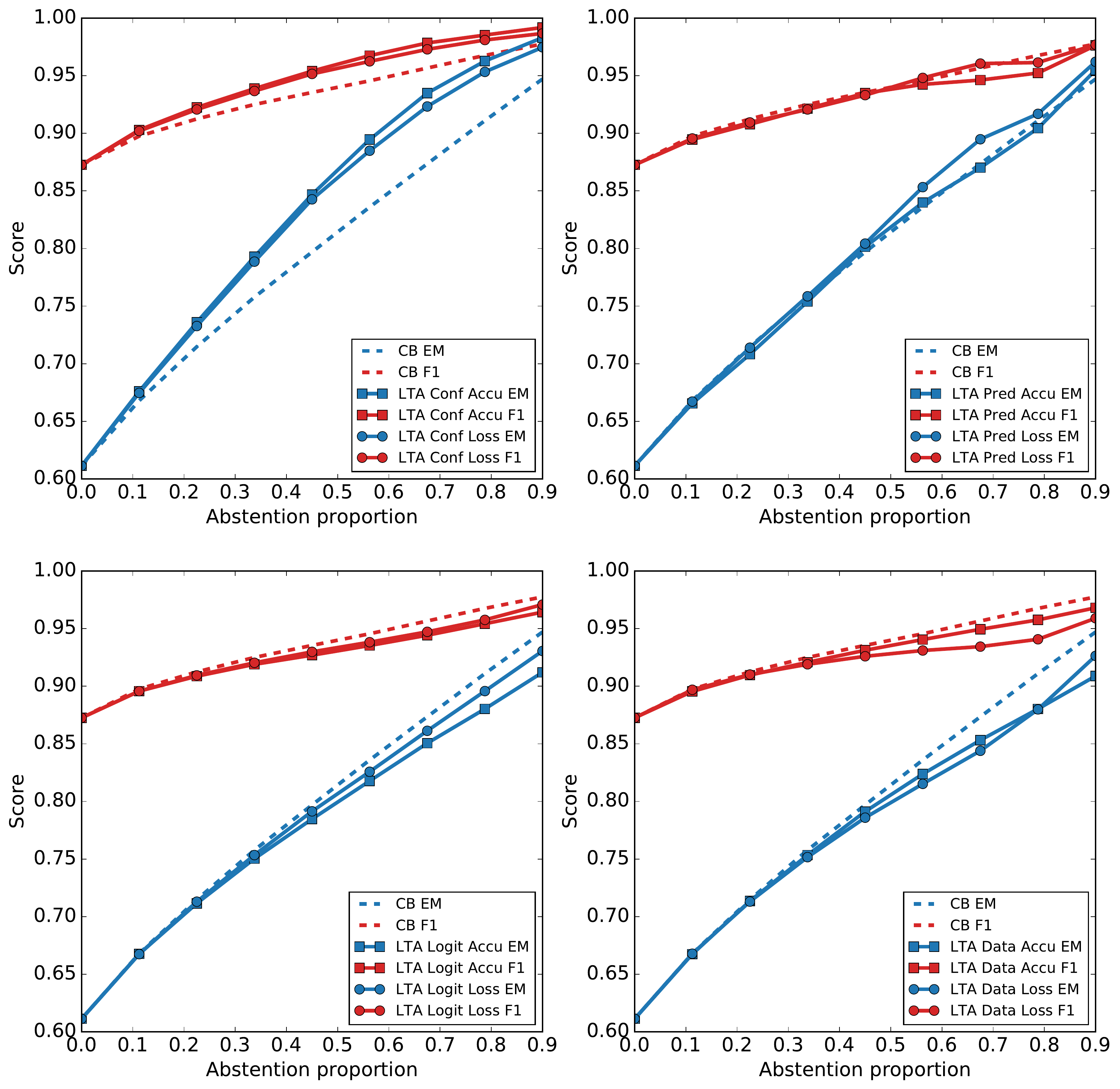}
\caption{{\bf Abstention improvement curve.} 
Top-left: learning to reject model with confidence score as input, estimate accuracy or loss. Top-right: learning to reject model with post-sigmoid probabilities $\bm{\hat y}$ score as input, estimate accuracy or loss. Bottom-left: learning to reject model with prior-to-sigmoid logits as input, estimate accuracy or loss. Bottom-right: learning to reject model with global max pooled hidden states $\bm{c}$ as input, estimate accuracy or loss.
}
\label{fig:abs-improv-curve}
\end{figure*}

The score $\alpha_{\text{conf}}$ is an abstention priority score because it is a valid indication of how confident the model's overall output is. We refer to this scheme confidence-based abstention module (or ``CB'' in Figure~\ref{fig:abs-improv-curve}, ``Baseline'' in Figure~\ref{fig:abs-best-curve}).

\paragraph{Learning-based abstention}

Instead of computing $\alpha$ from a fixed formula, we can try to link abstention priority score to a value that we care about. For example, we want to drop examples that will induce high loss, or equivalently, examples where predicted result gives a low accuracy. However, we do not have access to ground-truth answers in the real world, instead, we propose that if the data distribution $\mathcal{D}$ between training and deployment are consistent ($x_{\text{test}}, y_{\text{test}} \sim \mathcal{D}$, which is the underlying assumption specified in calibration), then we can learn to estimate loss or accuracy for each example. 
We can compute a regression target for the learned abstention module using the training dataset's accuracy and loss value for each example (Eq~\ref{eq:abs-true-obj}).
% In Eq~\ref{eq:abs-true-obj}, we show that we can compute document $i$'s accuracy $\alpha_{\text{accu}}^i$ or loss $\alpha_{\text{loss}}^i$ in our training dataset.

% estimate a meaningful abstention priority score given the true value. The idea of abstaining an example is based on the fact that the model will either get a high loss or a low accuracy if it chooses to predict the labels on such example. This makes accuracy or loss the perfect candidate for the abstention priority score. We can compute instance-based accuracy or average binary cross entropy loss as follow:

\begin{equation}
\begin{aligned}
\alpha_{\text{accu}}^i &= \frac{1}{m} \sum_{j=1}^m I(d(\bm{\hat y})^i_j, \bm{y}^i_j) \\
\alpha_{\text{loss}}^i &= \mathcal{L}_{\text{BCE}}(\bm{\hat y}^i, \bm{y}^i) \\
\end{aligned}
\label{eq:abs-true-obj}
\end{equation}

% , which can be either loss or accuracy computed above
This abstention learning module $A$ can take an input $z$ and output an estimated abstention score $\hat \alpha$. We train this module by minimizing minimum square squared error with the regression target:

\begin{equation}
\begin{aligned}
\hat \alpha^i &= A(z^i) \\ % {\text{learn}}
\mathcal{L}_{\text{MSE}} &= \sum_{i=1}^N (\alpha^i - \hat \alpha^i)^2 \\ % _{\text{learn}}
\end{aligned}
\label{eq:abs-learned}
\end{equation}

% Allen: below is not yet edited

We choose four possible inputs from various parts of the DeepTag model that the DeepTag-abstention module can use to predict accuracy or loss without knowing the ground-truth label. Two choices are obvious: confidence scores $g(\bm{\hat y})$ that is used to compute confidence-based abstention priority score in the previous section, and estimated probability for the presence of each label $\bm{\hat y}$, which we have used to compute confidence scores via function $g(\cdot)$. 
However, since $\bm{\hat y}$ is obtained by applying a sigmoid function to the output of the classifier $\hat y_i = \sigma(\theta_i^\intercal \bm{c})$, then we can also use the prior-to-sigmoid value $\theta_i^\intercal \bm{c}$ as input. At last, we hypothesize that the representation of document $\bm{c}$ might also contain relevant information that is useful for model $A$ to determine whether the document is difficult to process. 

% In order to find the appropriate input $z$ for the model $A$, we experiment with four different inputs: representation of the document: $\bm{c}$, prior-to-sigmoid logits: $\Theta \bm{c}$, post-sigmoid probabilities $\bm{\hat y}$, and confidence scores $g(\bm{\hat y})$.

We fit the model $A$ to estimate $\alpha_\text{learn}$ in the training set of our data, same split as the one used to train the overall model. We then evaluate on a previously unseen test set.

\subsection*{Experimental Details}

We initialize our model with 100-dimension GloVE word vectors~\cite{pennington2014glove}, and we initialize un-matched words in the CSU training data with sampled multivariate normally distributed vectors. We allow all word embeddings to be updated through the training process. We use a recurrent neural network with a 512 dimension LSTM cell, and set the feed-forward dropout rate to be 20\%. We use batch size of 32, clipping gradient at 5. We use ADAM \cite{kingma2014adam} optimizer with a learning rate of 0.001. 

% \edit{For the CNN baseline experiment, we control the output dimension of the CNN network to be exactly the same as BLSTM, which is 1024 dimension. We convolve on the word level with kernel size equal to 3}.

We trained all models to the maximum of 5 epochs with early stopping, the maximum number of epoch is picked by observing performance on validation dataset. After picking out the best hyper-parameters on validation set,  we evaluate all models in-domain generalization performance on the CSU test dataset and out-domain generalization performance on the PP dataset.

After hyperparameter searching, we are report models with the hyperparameters that perform well on each dataset. We train each model five times and report the averaged result.
For the CSU dataset, we find $\beta=0.001$ works best for DeepTag-M, and $\gamma_\text{norm}=1e-5, \gamma_\text{between}=1e-4, \gamma_\text{within}=1e-4$ works best for DeepTag (cluster penalty).
For the PP dataset, we find $\beta=0.0001$ works the best for DeepTag-M, and $\gamma_\text{norm}=1e-4, \gamma_\text{between}=1e-3, \gamma_\text{within}=1e-3$ works best for DeepTag (cluster penalty). We report these results in Table~\ref{tab:pp}.

For Table~\ref{tab:detailed-csu-pp}, we report DeepTag trained with  $\gamma_\text{norm}=1e-4, \gamma_\text{between}=1e-3, \gamma_\text{within}=1e-3$ and we regard this as our best setting.

\subsection*{Abstention Experimental Details}

We use a 3-layer neural network with SELU activation \cite{klambauer2017self} to parameterize abstention model $A$. The learning to abstain model is trained on various outputs generated by the DeepTag system after training the bidirectional LSTM with cluster penalty. All configurations of learning to abstain models are trained optimally for 3 epochs on the training set, and evaluated on the unseen test set.

% We use a 3-layer neural network with SELU activation \cite{klambauer2017self} to parameterize abstention model $A$. The learning to abstain model is trained after the basic LSTM classifier (without any loss augmentation) finished training. All configurations of learning to abstain models are trained optimally for 3 epochs on the training set, and evaluated on the unseen test set.

% \begin{figure}[h]
% \includegraphics[scale=0.4]{images/number-of-labels-abstained.png}
% \caption{{\bf Proportion of documents abstained and their label distribution}}
% \label{fig:number-of-labels}
% \label{fig:abstention-results}
% \end{figure}
% \james{This should be labeled as Supplementary Figure X.}

\begin{figure*}[!h]
\centering
\includegraphics[scale=0.35]{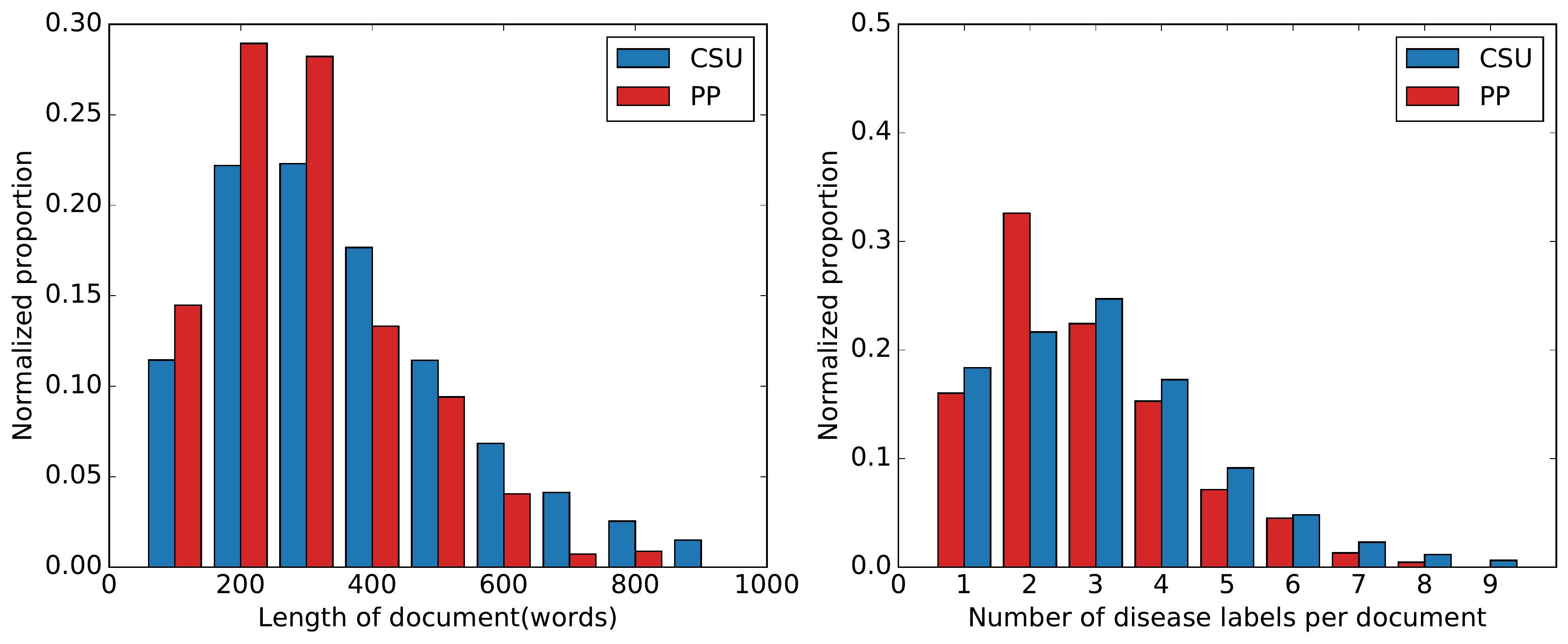}
\caption{{\bf Document length and label distribution on CSU and PP dataset.} 
Proportion of records in each dataset with certain length (number of words) or certain number of labels.}
\label{fig:doc-lengths-label-dist}
\end{figure*}

\subsection*{SNOMED Meta-categories}

Here we provide the full list of the SNOMED-CT meta-categories that we used to regularize the training objective of DeepTag. In the list, the numbers correspond to the meta-categories, and the letters indicate the original SNOMED-CT codes. We manually clustered the 42 SNOMED-CT codes into these 19 meta-categories, using the analogous clustering of the ICD-9 categories as a guide. 

\begin{enumerate}
\item Complications of pregnancy, childbirth, and the puerperium 
 \begin{enumerate} 
 \item Disorder of labor / delivery (disorder) 
 \item Disorder of pregnancy (disorder)
 \item Disorder of puerperium (disorder)
 \end{enumerate} 
\item Diseases of the genitourinary system 
 \begin{enumerate} 
 \item Disorder of the genitourinary system (disorder) 
 \end{enumerate} 
\item Diseases of the musculoskeletal system and connective tissue 
 \begin{enumerate} 
 \item Disorder of connective tissue (disorder) 
 \item Disorder of musculoskeletal system (disorder) 
 \end{enumerate} 
\item Diseases of the skin and subcutaneous tissue 
 \begin{enumerate} 
 \item Angioedema and/or urticaria (disorder) 
 \item Disorder of pigmentation (disorder) 
 \item Disorder of integument (disorder) 
 \end{enumerate} 
\item Certain conditions originating in the perinatal period 
 \begin{enumerate} 
 \item Disorder of fetus or newborn (disorder) 
 \end{enumerate} 
\item Congenital anomalies 
 \begin{enumerate} 
 \item Hereditary disease (disorder) 
 \item Congenital disease (disorder) 
 \item Familial disease (disorder)
 \end{enumerate} 
\item Injury and poisoning 
 \begin{enumerate} 
 \item Disorder caused by exposure to ionizing radiation (disorder) 
 \item Poisoning (disorder) 
 \item Traumatic AND/OR non-traumatic injury (disorder) 
 \item Self-induced disease (disorder)
 \end{enumerate} 
\item Symptoms, signs, and ill-defined conditions 
 \begin{enumerate} 
 \item Hyperproteinemia (disorder) 
 \item Clinical finding (finding) 
 \end{enumerate} 
\item Neoplasms 
 \begin{enumerate} 
 \item Neoplasm and/or hamartoma (disorder) 
 \item Fibromatosis (disorder)
 \end{enumerate} 
\item Infectious and parasitic diseases 
 \begin{enumerate} 
 \item Disease caused by Arthropod (disorder) 
 \item Disease caused by Annelida (disorder)
 \item Infectious disease (disorder)
 \item Disease of presumed infectious origin (disorder)
 \item Disease caused by parasite (disorder) 
 \item Enzootic disease (disorder)
 \item Epizootic disease (disorder)
 \end{enumerate} 
\item Diseases of blood and blood-forming organs 
 \begin{enumerate} 
 \item Anemia (disorder) 
 \item Disorder of cellular component of blood (disorder) 
 \item Disorder of hematopoietic cell proliferation (disorder) 
 \item Disorder of hemostatic system (disorder) 
 \item Spontaneous hemorrhage (disorder) 
 \item Hyperviscosity syndrome (disorder)
 \item Secondary and recurrent hemorrhage (disorder)
 \item Secondary hemorrhage (disorder)
 \end{enumerate} 
\item Endocrine, nutritional and metabolic diseses, and immunity disorders 
 \begin{enumerate} 
 \item Autoimmune disease (disorder) 
 \item Disorder of immune function (disorder) 
 \item Hypersensitivity condition (disorder) 
 \item Metabolic disease (disorder) 
 \item Nutritional deficiency associated condition (disorder) 
 \item Nutritional disorder (disorder) 
 \item Obesity (disorder)
 \item Obesity associated disorder (disorder)
 \item Propensity to adverse reactions (disorder) 
 \item Disorder of endocrine system (disorder) 
 \end{enumerate} 
\item Diseases of the nervous system 
\item Feline hyperesthesia syndrome (disorder)
 \begin{enumerate} 
 \item Disorder of nervous system (disorder) 
 \end{enumerate} 
\item Mental disorders 
 \begin{enumerate} 
 \item Mental disorder (disorder) 
 \end{enumerate} 
\item Diseases of the circulatory system 
 \begin{enumerate} 
 \item Disorder of cardiovascular system (disorder) 
 \end{enumerate} 
\item Diseases of sense organs 
 \begin{enumerate} 
 \item Disorder of auditory system (disorder) 
 \item Vertiginous syndrome (disorder)
 \item Visual system disorder (disorder) 
 \item Sensory disorder (disorder)
 \end{enumerate} 
\item Diseases of the digestive system 
 \begin{enumerate} 
 \item Vomiting (disorder)
 \item Enterotoxemia (disorder)
 \item Disorder of digestive system (disorder) 
 \end{enumerate} 
\item Diseases of the respiratory system 
 \begin{enumerate} 
 \item Disorder of respiratory system (disorder) 
 \end{enumerate} 
\end{enumerate}

\end{document}